\documentclass[acmtog]{acmart}
\AtBeginDocument{%
  }

\copyrightyear{2026}
\acmYear{2026}
\setcopyright{cc}
\setcctype{by}
\acmConference[SIGGRAPH Conference Papers '26]{Special Interest Group on Computer Graphics and Interactive Techniques Conference Conference Papers}{July 19--23, 2026}{Los Angeles, CA, USA}
\acmBooktitle{Special Interest Group on Computer Graphics and Interactive Techniques Conference Conference Papers (SIGGRAPH Conference Papers '26), July 19--23, 2026, Los Angeles, CA, USA}
\acmDOI{10.1145/3799902.3811126}
\acmISBN{979-8-4007-2554-8/2026/07}

\acmSubmissionID{732}

\usepackage{color, soul}

\soulregister{\cite}{7}
\soulregister{\ref}{7}
\soulregister{\pageref}{7}

\usepackage{amsthm}  
\usepackage{multirow}
\usepackage{enumitem}
\usepackage{appendix}
\usepackage[capitalise]{cleveref}
\AtBeginEnvironment{appendices}{\crefalias{section}{appendix}}
\usepackage{xcolor} 
\usepackage[linesnumbered,ruled,vlined]{algorithm2e}

\definecolor{soft_red}{RGB}{238,124,146}
\definecolor{soft_blue}{RGB}{114,161,219}

\begin{document}

\title{VecSet-Edit: Unleashing Pre-trained LRM for Mesh Editing from Single Image}

\author{Teng-Fang Hsiao}
\author{Bo-Kai Ruan}
\author{Yu-Lun Liu}
\author{Hong-Han Shuai}

\affiliation{%
  \institution{National Yang Ming Chiao Tung University}
  \country{Taiwan}
}

\newcommand{\ie}{\textit{i.e.}, }
\newcommand{\eg}{\textit{e.g.}, }
\begin{abstract}
3D editing has emerged as a critical research area to provide users with flexible control over 3D assets. While current editing approaches predominantly focus on 3D Gaussian Splatting or multi-view images, the direct editing of 3D meshes remains underexplored. Prior attempts, such as VoxHammer, rely on voxel-based representations that suffer from limited resolution and necessitate labor-intensive 3D masks. To address these limitations, we propose \textbf{VecSet-Edit}, the first pipeline that leverages the high-fidelity VecSet Large Reconstruction Model (LRM) as a backbone for mesh editing. Our approach is grounded on an analysis of the spatial properties in VecSet tokens, revealing that token subsets govern distinct geometric regions. Based on this insight, we introduce \textbf{Mask-guided Token Seeding} and \textbf{Attention-aligned Token Gating} strategies to precisely localize target regions using only 2D image conditions. Also, considering the difference between VecSet diffusion process versus voxel-based ones, we design a \textbf{Drift-aware Token Pruning} to reject geometric outliers during the denoising process. Finally, our \textbf{Detail-preserving Texture Baking} module ensures that we not only preserve the geometric details of the original mesh but also the textural information. Our project can be found in \url{https://github.com/BlueDyee/VecSet-Edit}
\end{abstract}


\begin{teaserfigure}
  \includegraphics[width=\textwidth]{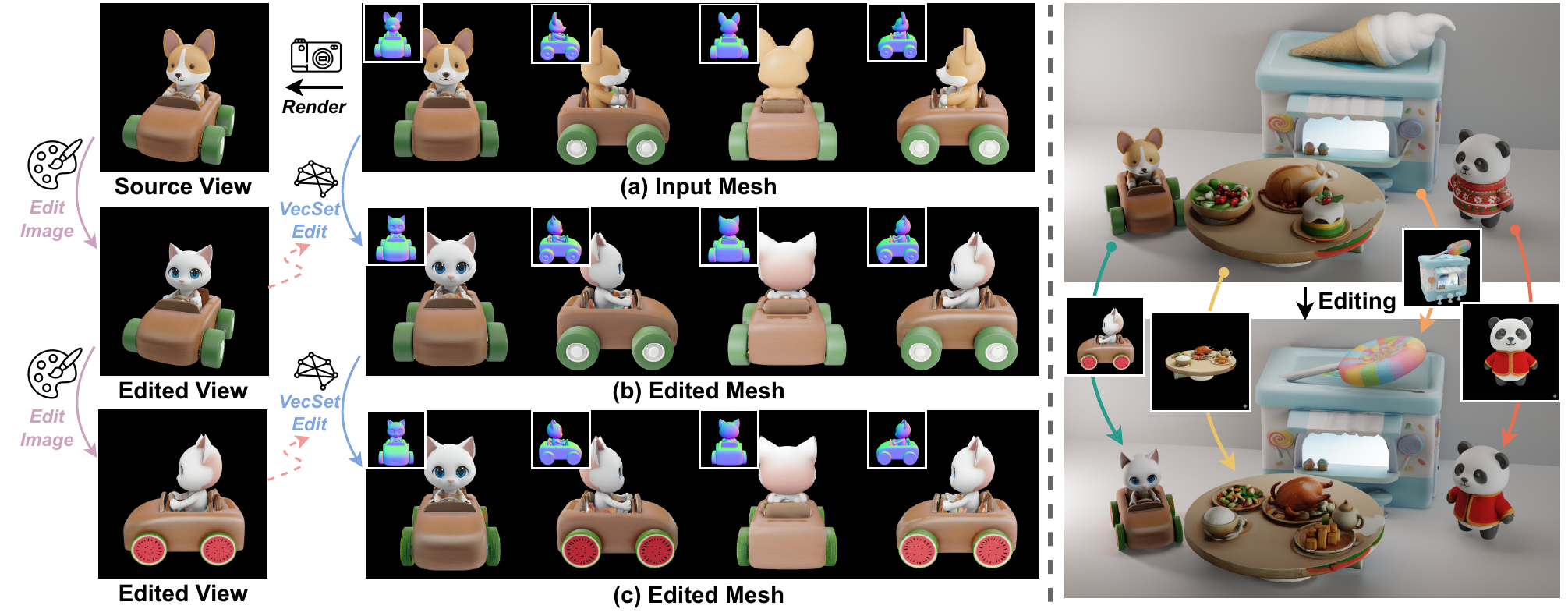}
  \vspace{-18pt}
\caption{
\textbf{VecSet-Edit: Localized geometry and texture editing from a single image.}
Our method allows for localized 3D mesh editing guided by a single-view 2D image. (\emph{Left}) Given an input mesh (a), users can edit a rendered view to guide the 3D editing. As shown in (b) and (c), our method accurately transfers these 2D semantic changes to the 3D mesh. This produces explicit geometric deformations (e.g., cat head, watermelon wheels) while preserving the geometry of unedited regions (e.g., the car body remains unchanged). (\emph{Right}) This localized capability enables precise multi-object editing in complex scenes.
}
  \label{fig:teaser}
\end{teaserfigure}


\maketitle
\section{Introduction}
Recent progress in automated 3D asset generation~\cite{trellis,dora_autoencoder,craftsman3d,hunyuan3d20,triposg} is rapidly increasing the supply of 3D content. Yet in real production workflows, generated assets rarely satisfy downstream requirements. Creators need localized, controllable refinements, such as editing a specific part, adjusting geometry in a region, or modifying appearance while preserving the asset identity. Re-generation is typically unreliable, because it can inadvertently change geometry or texture, making 3D editing essential for turning generated outputs into usable assets.

Despite this need, most existing editing methods operate on intermediate representations such as 3D Gaussian Splatting (3DGS)~\cite{gaussianeditor_chen,gaussianeditor_wang,3dsceneeditor,splatpainter,painting_3d_gaussian} or multi-view images~\cite{cmd,tip_editor,editp23,sharpit,edit360}, rather than directly editing meshes with explicit topology required by animation and physics-based simulation. While VoxHammer~\cite{voxhammer} takes an important step by leveraging a voxel-based Large Reconstruction Model (LRM)~\cite{trellis} for editing, it still faces two practical limitations: 1) it requires additional 3D mask annotation that is labor intensive and hard to scale, and 2) its voxel granularity fundamentally constrains resolution and fidelity compared with modern VecSet-based LRMs. These gaps call for a mesh native editing framework that demands only lightweight supervision, preserves asset identity, and fully exploits high fidelity VecSet reconstruction backbones.

To address these limitations, we propose \textbf{VecSet-Edit}, a training-free framework for localized 3D mesh editing built on a high-fidelity VecSet reconstruction model~\cite{clay,craftsman3d,triposg,hunyuan3d_omni,hunyuan3d20,step1x}. Given a reference mesh and its rendered view, a target edited image, and a binary 2D mask, VecSet-Edit produces an edited mesh that follows the requested change while preserving identity and fine details outside the edit region (as shown in \cref{fig:teaser}). The core challenge is localization: VecSet encodes geometry as an unordered token set, so naive token-space editing often affects unintended regions. \textit{Our key observation is that VecSet tokens are not spatially arbitrary; despite the unordered form, they exhibit stable locality and consistently correspond to coherent surface regions.} Leveraging this property, we localize edits by operating on a region-specific token subset. Since 2D-to-token selection can be noisy, we introduce \textbf{Mask-guided Token Seeding} to obtain a coarse token set and \textbf{Attention-aligned Token Gating} to retain only tokens most correlated with the target region, tightening localization and preventing spillover.

With localization in place, we face a second challenge unique to VecSet diffusion. During denoising, tokens are spatially mobile rather than fixed on a grid, and early drift can cause irreversible interference between edited and preserved regions. VecSet-Edit addresses this with \textbf{Drift-aware Token Pruning}, which removes drifted tokens that violate geometric consistency with the reference, keeping boundaries clean. Finally, \textbf{Detail-preserving Texture Baking} transfers texture to the edited mesh while retaining high-frequency details from the original asset. The key contributions can be summarized as follows. 
\begin{itemize}[leftmargin=*]
\item We present \textbf{VecSet-Edit}, the first training-free framework that enables localized 3D mesh editing directly in the latent VecSet space of a high-fidelity reconstruction model, avoiding the resolution bottleneck of voxel-based LRMs.
\item To localize edits on an unordered set of VecSet tokens, \textbf{Mask-guided Token Seeding} first derives a coarse editable token set from a 2D mask, and \textbf{Attention-aligned Token Gating} then retains tokens with the strongest spatial correlation to the target region. This ensures editing does not leak to unintended areas.
\item We further introduce \textbf{Drift-aware Token Pruning} to remove drifted tokens that would otherwise break geometric consistency during the denoising process, and propose \textbf{Detail-preserving Texture Baking} to update textures only where geometry changes, preserving appearance details in untouched regions.
\end{itemize}

\section{Related Works}
\subsection{VecSet-based Large Reconstruction Models}
Recent progress in Large Reconstruction Models (LRMs) has enabled high-quality 3D asset synthesis from text prompts or images. Existing approaches can be grouped by their intermediate 3D representations, including voxel-based methods~\cite{trellis,hunyuan3d_10,CRM}, triplane-based methods~\cite{masked_lrm,shap-e,LRM_stylization}, and VecSet-based methods~\cite{craftsman3d,triposg,hunyuan3d_omni,hunyuan3d20,step1x}. While voxel and triplane representations provide convenient spatial parameterizations, they typically rely on grid-aligned feature fields whose decoded surfaces do not necessarily expose mesh-level granularity for region-targeted edits. In contrast, VecSet-based LRMs represent surface geometry with a latent vector set, and have recently shown strong reconstruction quality with practical inference efficiency. However, their lack of explicit spatial indexing makes it unclear how to associate user-specified edit regions with a subset of latent tokens, which is crucial for localized mesh editing with identity preservation. This gap motivates editing frameworks that uncover reliable token to surface correspondence and support region-aware manipulation on top of VecSet generative backbones.

\subsection{3D Editing across Representations}
Neural 3D editing has moved beyond classical mesh deformation toward methods that modify learned scene representations, including NeRF-based models~\cite{instruct_nerf2nerf,vica_nerf,dreameditor} and 3D Gaussian Splatting~\cite{gaussianeditor_chen,gaussianeditor_wang,gaussctrl,stylizedgs,drag_gauss}. For instance, Instruct-NeRF2NeRF~\cite{instruct_nerf2nerf} applies diffusion-based guidance on rendered views and iteratively updates the underlying NeRF to realize the requested edits. Despite strong visual quality, these approaches primarily operate in radiance-field or splat space, making it nontrivial to obtain a mesh.

CMD~\cite{cmd} shares the same high-level goal as our method by focusing on 3D object editing via a single image. However, CMD relies on multi-view rendering and editing, which inevitably leads to inherent self-occlusions and resolution-bound reconstruction losses. On the other hand, LRM-based mesh editing methods~\cite{3dpaint_brush,vox_e,steer3d,ye2025nano3d} have shown promising growth. For example, VoxHammer~\cite{voxhammer} moves closer to native mesh editing via a voxel-based LRM, but it faces a fidelity-cost tension at high resolutions and heavily depends on explicit 3D region specifications. Conversely, by exploiting the compact yet expressive VecSet structure, our method delivers high-fidelity mesh editing results using only a 2D mask condition, bridging the gap between generation and precise control.

\section{Preliminary}
\label{sec:preliminary}

We build upon TripoSG~\cite{triposg}, a \emph{VecSet}-based LRM representing geometry as a compact set of latent vectors. It consists of a Variational Autoencoder (VAE) for mesh-token mapping and a Diffusion Transformer (DiT) for conditional generation. We leverage the VAE to bridge mesh and latent spaces, enabling localized editing by manipulating a selected subset of tokens during DiT denoising. See Appendix A for backbone details.

\smallskip

\begin{figure*}[t]
\includegraphics[width=\textwidth]{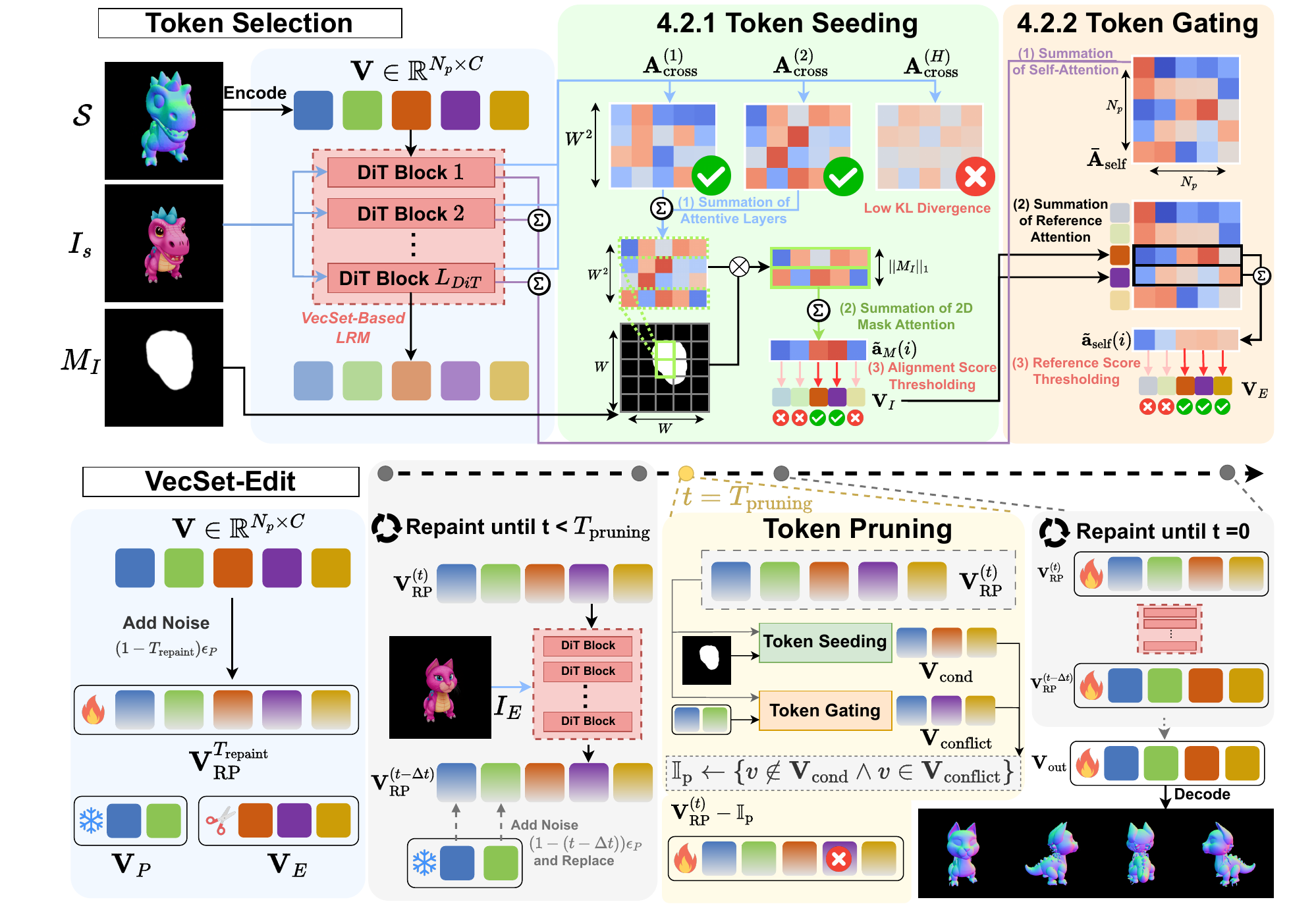}
\caption{\textbf{Overview of the VecSet-Edit framework.} Given a mesh $\mathcal{S}$, a rendered view $I_s$, a 2D edit mask $M_I$, and a user-edited target view $I_E$, the pipeline proceeds in two main stages. \textbf{First, Token Selection}: To localize the editable region without 3D supervision, \emph{Token Seeding} aggregates informative cross-attention layers to identify initial seed tokens $\mathbf{V}_I$ that align with the 2D mask. \emph{Token Gating} then leverages self-attention correlations to expand this selection to the full geometric structure, yielding the final editable subset $\mathbf{V}_E$. \textbf{Second, VecSet-Edit}: We perform diffusion-based editing on $\mathbf{V}_E$ while constraining the preserved tokens $\mathbf{V}_P$. To prevent geometric artifacts, \emph{Token Pruning} is applied during denoising to detect and discard ``conflict'' tokens that drift into the preserved regions without support from the editing condition. This ensures the final output faithfully respects both the target edit and the original structure.
}
\label{fig:pipeline}
\end{figure*}

\paragraph{VecSet VAE (geometry codec).} The VAE maps between a surface mesh and its latent representation. The encoder takes a surface point cloud $\mathbf{P}' \in \mathbb{R}^{N_{p'}\times 6}$ and learnable queries $\mathbf{P} \in \mathbb{R}^{N_{p}\times 6}$ (containing position and normal). Queries extract geometry from $\mathbf{P}'$ via cross-attention to yield latent tokens $\mathbf{V} = \operatorname{Encode}(\mathcal{S}) \in \mathbb{R}^{N_{p} \times C}$. Conversely, the decoder predicts a signed distance field (SDF) from $\mathbf{V}$ to extract the reconstructed mesh $\mathcal{S}' = \operatorname{Decode}(\mathbf{V})$ via Marching Cubes. We use these interfaces to seamlessly transition between mesh and VecSet spaces.


\paragraph{VecSet DiT (conditional token denoising).} The DiT performs conditional generation over latent tokens via rectified flow~\cite{flow}. Given an image $I \in \mathbb{R}^{H \times W \times 3}$, we extract dense features $h_I \in \mathbb{R}^{H \times W \times D}$ using a frozen encoder (e.g., DINOv2~\cite{dinov2}). Inference starts from noise $\mathbf{V}_T \sim \mathcal{N}(0, \mathbf{I})$ and iteratively evolves toward $\mathbf{V}_0$ using an Euler solver:
\begin{equation} \small
    \mathbf{V}_{t-\Delta t} = \mathbf{V}_t - u_{\theta}(\mathbf{V}_t, h_I, t)\cdot \Delta t,
\end{equation}
where $u_{\theta}$ is the predicted velocity field, $t$ is diffusion time, and $\Delta t=1/N_s$ is the step size.

The DiT is composed of $L_{\text{DiT}}$ stacked blocks that alternate between self-attention over VecSet tokens and cross-attention to image features. For the $l$-th block, the update is given by:
\begin{equation} \small
    \widehat{\mathbf{V}}^{(l)}_t = \operatorname{SelfAttn}(\mathbf{V}^{(l)}_t, t), \label{eq:self_attn}
\end{equation}
\begin{equation} \small
    \mathbf{V}^{(l+1)}_t = \operatorname{CrossAttn}(\widehat{\mathbf{V}}^{(l)}_t, h_I),\label{eq:cross_attn}
\end{equation}
where feed-forward layers and residual connections are omitted for clarity. These attention operations form the foundation of our token-level analysis and manipulation, which we exploit for localized mesh editing in the following sections.

\section{Method}
\label{sec:method}

We consider the task of 3D mesh editing guided by 2D image conditions. The overall pipeline is illustrated in \cref{fig:pipeline}. The input consists of a reference mesh $\mathcal{S}$ and its rendered view $I_{\mathcal{S}}\in\mathbb{R}^{H\times W\times 3}$, a target edited image $I_E\in\mathbb{R}^{H\times W\times 3}$ and a binary mask $M_I\in\mathbb{R}^{H\times W}$ indicating the region to be edited in the image space. Our goal is to (1) produce an edited mesh $\mathcal{S}_\text{out}$ whose geometry reflects the semantic changes specified by $I_E$ within the masked region, and (2) faithfully preserve the original geometry and texture of $\mathcal{S}$ elsewhere. To achieve this goal, we first leverage the \emph{VecSet Geometry Property} (\cref{sec:vecset_geometry}), which reveals that localized mesh regions can be controlled through appropriate subsets of VecSet tokens. This observation allows us to reformulate editing as a token selection problem. Building on this formulation, we introduce \textit{Token Selection} (\cref{sec:token_selection}) to map the 2D mask to geometry-relevant tokens. Based on the selected tokens, we propose \textbf{VecSet-Edit} (\cref{sec:vecset_edit}), a unified pipeline that performs diffusion-based VecSet editing.

\subsection{VecSet Geometry Property}
\label{sec:vecset_geometry}

Editing a mesh through its VecSet representation differs fundamentally from \textit{de novo} generation: we must preserve existing geometry in protected regions while modifying others. However, VecSet tokens lack an explicit spatial partition, and attention creates non-local dependencies, so a token can affect multiple regions rather than a single isolated part. Therefore, controllable local editing hinges on a concrete token–region association: \textit{Can we identify the tokens that most strongly influence a user-specified region, enabling targeted edits while keeping the rest unchanged?}

\begin{figure}[ht]
\centering
\includegraphics[width=\columnwidth]{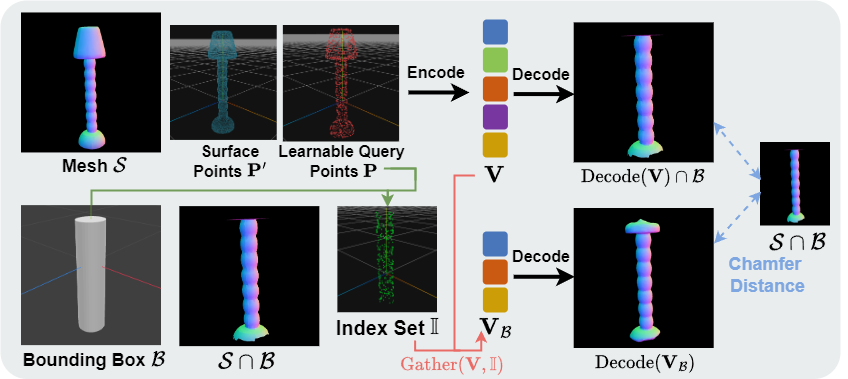}
\caption{
\textbf{Illustration of the VecSet Geometry Property.} 
We validate that the unordered VecSet tokens exhibit spatial locality. Given a mesh $\mathcal{S}$ and a bounding box $\mathcal{B}$, we first identify the index set $\mathbb{I}$ corresponding to query points $\mathbf{P}$ that fall within $\mathcal{B}$. We then extract the token subset $\mathbf{V}_{\mathcal{B}}=\text{Gather}(\mathbf{V}, \mathbb{I})$. Finally, we quantify the reconstruction fidelity by measuring the Chamfer Distance between the geometry decoded purely from the subset ($\text{Decode}(\mathbf{V}_{\mathcal{B}})$), the reference geometry cropped from the full reconstruction ($\operatorname{Decode}(\mathbf{V}) \cap \mathcal{B}$) with cropped source mesh $\mathcal{S}\cap \mathcal{B}$.
}
\label{fig:geometry_property}
\end{figure}

To answer this, we first formalize region-faithful token subsets using a 3D bounding volume and a reconstruction tolerance.
{
\setlength{\parindent}{0pt}
\begin{definition}[\textbf{VecSet Geometry Property}]
\label{def:geometry_property}
Given a 3D watertight bounding volume $\mathcal{B}$ and a mesh $\mathcal{S}$, we define the target region as their intersection, $\mathcal{S}_{\mathcal{B}} \coloneq \mathcal{S} \cap \mathcal{B}$. We say that a subset of VecSet tokens $\mathbf{V}_{\mathcal{B}} \subset \mathbf{V}$ satisfies the \textbf{geometry property} for $\mathcal{S}_{\mathcal{B}}$ under tolerance $\epsilon$ if the Chamfer Distance (CD) between $\mathcal{S}_{\mathcal{B}}$ and the geometry decoded from $\mathbf{V}_{\mathcal{B}}$ is below $\epsilon$:
\begin{equation} \small
    \operatorname{CD}(\mathcal{S}_{\mathcal{B}}, \operatorname{Decode}(\mathbf{V}_{\mathcal{B}})) < \epsilon.
\end{equation}
\end{definition}
}

\paragraph{Empirical verification.} We conduct a sanity check with TripoSG~\cite{triposg} on Edit3D-Bench~\cite{voxhammer} (300 pairs of mesh and bounding box). To identify the subset $\mathbf{V}_\mathcal{B}$ corresponding to the edit region, we leverage the spatial alignment inherent in the VecSet encoding (Appendix A.1). Recall that each token is initialized via positional encodings derived from a query point $\mathbf{p}_i \in \mathbb{R}^{6}$. 
Based on the premise that these tokens retain strong spatial priors from the encoding process, we select the subset falling within the box $\mathcal{B}$. 
Formally, we define the index set $\mathbb{I} = \{ i \mid \mathbf{p}_i^{\text{xyz}} \in \mathcal{B} \}$, where $\mathbf{p}_i^{\text{xyz}}$ denotes the spatial coordinates of $\mathbf{p}_i$. The local VecSet representation is then constructed by restricting the latent to this subset: $\mathbf{V}_\mathcal{B} = \{ \mathbf{v}_i \}_{i \in \mathbb{I}}$. The illustration of this verification pipeline can be found in \cref{fig:geometry_property}.

\smallskip

\begin{table}[t]
    \centering
    \small
    \caption{\textbf{Reconstruction fidelity of the VecSet subset versus $\mathcal{S}_\mathcal{B}$.} We report the percentage of samples with a Chamfer Distance smaller than the threshold $\epsilon$.}
    \begin{tabular}{lcccc}
      \toprule
      Subset Type & $\epsilon=0.30$& $\epsilon=0.10$& $\epsilon=0.05$& $\epsilon=0.01$\\
      \midrule
      $\operatorname{Decode}(\mathbf{V}) \cap \mathcal{B}$ & $100\%$& $100\%$ &$100\%$&$98.6\%$\\
      $\operatorname{Decode}(\mathbf{V}_\mathcal{B})$ & $82.3\%$& $73.5\%$ & $69.4\%$& $44.2\%$\\
     \bottomrule
    \end{tabular}
    \label{tab:vecset_geometry_property}
\end{table}

\paragraph{Results.} As shown in \cref{tab:vecset_geometry_property}, $82.3\%$ of samples reconstructed from the subset $\mathbf{V}_\mathcal{B}$ achieve a Chamfer Distance below $\epsilon=0.30$, while $44.2\%$ meet the stricter threshold of $\epsilon=0.01$. Although tight tolerances remain challenging, these results indicate that even this naive selection often yields a region-faithful subset, supporting the practical relevance of the \emph{VecSet Geometry Property}. We expect the reconstruction fidelity to further improve with more robust token selection or refinement strategies.


\paragraph{Connection with Editing.} The property motivates decoupled mesh editing by partitioning the latent tokens into an editable subset and a preservation subset: $\mathbf{V} \coloneq \mathbf{V}_E \oplus \mathbf{V}_P,$
where $\mathbf{V}_E$ is intended to capture the user-specified edit region and $\mathbf{V}_P$ preserves the remaining structure. Under this view, controllable editing reduces to a \emph{token selection} problem: identifying $\mathbf{V}_E$ for an instruction-defined target region while keeping $\mathbf{V}_P$ unchanged.

\subsection{Token Selection}\label{sec:token_selection}
Guided by the VecSet Geometry Property, we introduce a two \emph{Token Selection} mechanism to address distinct editing requirements. To ensure semantic alignment with the 2D condition, we employ \emph{Mask-guided Token Seeding}; to maintain 3D spatial coherence, we utilize \emph{Attention-aligned Token Gating}.

\subsubsection{Mask-guided Token Seeding}
\label{sec:image_selection}

Inspired by prior text-to-image editing works~\cite{prompt_to_prompt,tf_ti2i,tf-gph,diffedit,sdedit,multi_turn,kulikov2025flowedit,cao2023masactrl}, which demonstrate that cross-attention maps provide effective zero-shot cues for semantic localization, we adapt this idea to the 3D VecSet DiT to identify geometry-relevant tokens from a 2D edit mask.
Intuitively, if a VecSet token consistently attends to pixels inside the mask, it is likely responsible for generating geometry that explains the masked region.

Let $\mathbf{A}^{(l,t)}_{\text{cross}}\in\mathbb{R}^{N_p\times HW}$ denote the cross-attention map in the VecSet DiT at block $l$ and diffusion timestep $t$ (refer to~\cref{eq:cross_attn}), and let $\bar{M}_I\in\{0,1\}^{HW}$ be the flattened 2D mask. We first measure the alignment between the $i$-th token and the masked region by accumulating its attention mass over masked pixels:
\begin{equation} \small
    a^{(l,t)}_M(i)=\sum_{j=1}^{HW}\mathbf{A}^{(l,t)}_{\text{cross}}[i,j]\cdot \bar{M}_I[j].
\end{equation}

\begin{figure}
\includegraphics[width=0.9\columnwidth]{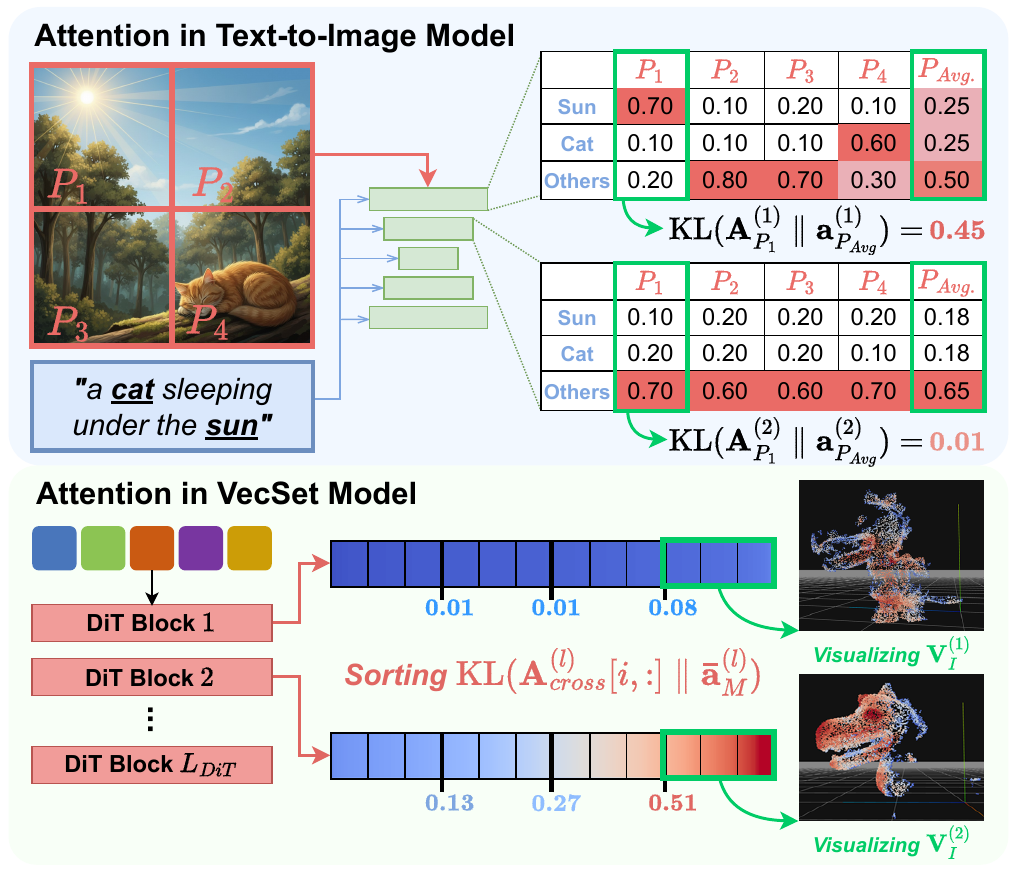}
\caption{\textbf{Illustration of KL divergence in T2I and VecSet Diffusion process.} In the T2I models, the layers with higher divergence are more correlated the prompt with object location. A similar pattern can be found in the VecSet Model, where the tokens with higher KL divergence are more correlated with the image.}\label{fig:kl}
\end{figure}

Since attention quality is inconsistent across layers~\cite{multi_turn}, naively aggregating all blocks introduces noise. Drawing on the intuition that high KL divergence correlates with strong condition alignment (as shown in ~\cref{fig:kl}), we evaluate layer informativeness via the KL-divergence of their cross-attention distributions. Consequently, we restrict aggregation to a subset of layers $\mathcal{L}_{\text{attn}}$, filtering out uninformative patterns (details in Appendix B), and average over timesteps:
\begin{equation} \small
    \tilde{a}_M(i)=\mathbb{E}_{l\in\mathcal{L}_{\text{attn}},\,t}\!\left[a^{(l,t)}_M(i)\right].
\end{equation}
We select tokens whose alignment score exceeds a threshold $\tau_I$\footnote{The choice of $\tau_I$ can be found in~Appendix~B.}:
\begin{equation} \small
    \mathbf{V}_I=\texttt{Seeding}(\mathbf{V},I_S,M_I)
    =\left\{\mathbf{v}_i\in\mathbf{V}\ \middle|\ \tilde{a}_M(i)>\tau_I\right\}.
\end{equation}
$\mathbf{V}_I$ represents the subset of VecSet tokens whose decoded geometry explains the masked region in the conditioned view. As such, $\mathbf{V}_I$ serves as a semantic proxy for the 3D edit region inferred purely from 2D image supervision.

\subsubsection{Attention-aligned Token Gating}
\label{sec:attention_selection}

To further improve spatial coherence, we leverage the self-attention maps in the VecSet DiT, which capture token–token interactions that often reflect geometric proximity and topological adjacency. Starting from a reference token set $V_I$, we use these attention-derived affinities to expand the set by retrieving additional tokens that are most strongly coupled to the same local region, yielding a more spatially consistent token subset for editing. Specifically, let $\mathbf{A}^{(l,t)}_{\text{self}}\in\mathbb{R}^{N_p\times N_p}$ denote the self-attention map at block $l$ and timestep $t$ (refer to \cref{eq:self_attn}). Given a reference token subset $\mathbf{V}_{\text{ref}}\subset\mathbf{V}$, we define its membership indicator $\mathbf{m}_{\mathbf{V}_{\text{ref}}}\in\{0,1\}^{N_p}$, where $\mathbf{m}_{\mathbf{V}_{\text{ref}}}[j]=1$ iff $\mathbf{v}_j\in\mathbf{V}_{\text{ref}}$. We then score each token by how much it attends to the reference set:
\begin{equation} \small
    a^{(l,t)}_{\text{self}}(i)=\sum_{j=1}^{N_p}\mathbf{A}^{(l,t)}_{\text{self}}[i,j]\cdot \mathbf{m}_{\mathbf{V}_{\text{ref}}}[j].
\end{equation}
Unlike cross-attention, self-attention exhibits more stable query-key relationships across layers. As such, we average uniformly:
\begin{equation} \small
    \tilde{a}_{\text{self}}(i)=\mathbb{E}_{l,\,t}\!\left[a^{(l,t)}_{\text{self}}(i)\right].
\end{equation}
Similarly, we select tokens whose score exceeds a threshold $\tau_A$:
\begin{equation} \small
    \mathbf{V}_A=\texttt{TokenGating}(\mathbf{V},\mathbf{V}_{\text{ref}})
    =\left\{\mathbf{v}_i\in\mathbf{V}\ \middle|\ \tilde{a}_{\text{self}}(i)>\tau_A\right\}.
\end{equation}
The resulting subset $\mathbf{V}_A$ effectively captures tokens geometrically adjacent to the reference set. This strategy leverages the geometry property of VecSets, where strong self-attention correlations serve as a reliable proxy for geometric connectivity. 


\subsection{VecSet-Edit Framework}
\label{sec:vecset_edit}

We now present \textbf{VecSet-Edit}, a framework for image-guided localized mesh editing in VecSet space. Given a reference mesh $\mathcal{S}$ encoded as VecSet tokens $\mathbf{V}$, a source image $I_S$, a target edited image $I_E$, and a binary edit mask $M_I$, the objective is to modify only the geometry associated with the masked region while strictly preserving the remaining structure. VecSet-Edit consists of three tightly coupled stages: (i) identifying editable tokens via token selection, (ii) performing constrained diffusion-based editing, and (iii) removing diffusion-induced artifacts through token pruning.


\subsubsection{Editable Token Decomposition. }
Following the VecSet Geometry Property (\cref{sec:vecset_geometry}), we first decompose the VecSet representation into editable and preserved subsets. To obtain $\mathbf{V}_E$, we apply Mask-guided Token Seeding using the source image and mask:
\begin{equation} \small
    \mathbf{V}_I = \texttt{TokenSeeding}(\mathbf{V}, I_S, M_I),
\end{equation}
which identifies tokens whose decoded geometry explains the masked image region. We then enforce spatial coherence by expanding this set through Attention-aligned Token Gating:
\begin{equation} \small
    \mathbf{V}_E = \texttt{TokenGating}(\mathbf{V}, \mathbf{V}_I),
\end{equation}
with the preserved tokens defined as $\mathbf{V}_P = \mathbf{V} \setminus \mathbf{V}_E$.


\subsubsection{RePaint in VecSet Space.}
With $\mathbf{V}_P$ fixed, we perform diffusion-based editing by adapting the RePaint strategy~\cite{lugmayr2022repaint} to VecSet tokens. Starting from a noisy initialization at timestep $T_{\text{repaint}}$, the editable tokens are iteratively denoised under the guidance of the target image $I_E$, while the preserved tokens are constrained to follow their original diffusion trajectory:
\begingroup
\small
\begin{align}
    \mathbf{V}_{E}^{(t-\Delta t)} &= \mathbf{V}_{E}^{(t)} - 
    \texttt{Gather}\!\left(u_{\theta}(\mathbf{V}_{\text{RP}}^{(t)}, h_I, t), \mathbb{I}_E\right)\cdot \Delta t, \\
    \mathbf{V}_{P}^{(t-\Delta t)} &= (1-(t-\Delta t))\,\epsilon_P + (t-\Delta t)\,\mathbf{V}_P, \\
    \mathbf{V}_{\text{RP}}^{(t-\Delta t)} &= \mathbf{V}_{E}^{(t-\Delta t)} \oplus \mathbf{V}_{P}^{(t-\Delta t)}.
    \label{eq:repaint}
\end{align}
\endgroup
Here, $\mathbb{I}_E$ denotes the set of indices corresponding to the editable tokens $\mathbf{V}_E$, and $\texttt{Gather}(\cdot, \mathbb{I}_E)$ extracts the output predictions from the backbone model based on these indices. This process allows the edited region to migrate toward the target geometry while maintaining consistency with the preserved structure.

\begin{figure}
\includegraphics[width=\columnwidth]{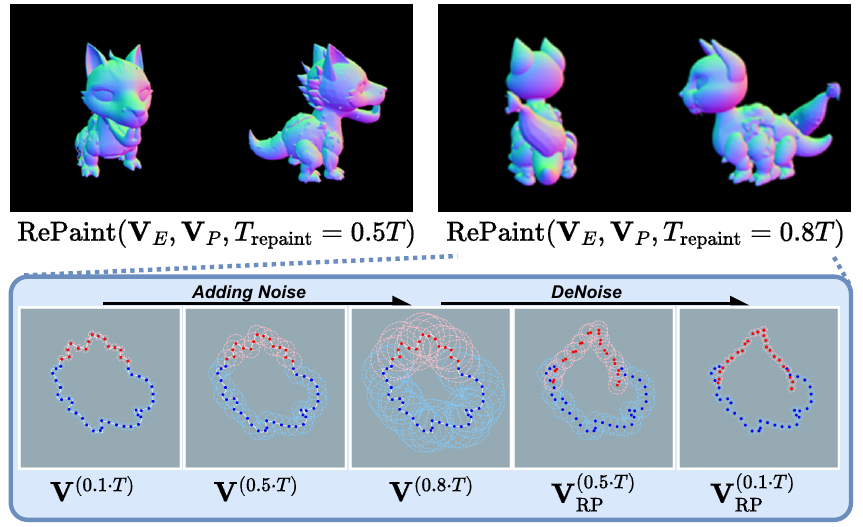}
\caption{\textbf{Illustration of the VecSet RePaint process (same input condition as \cref{fig:pipeline}).} We visualize a toy example where tokens serve as particles and their movement regions are denoted by circles. \textcolor{soft_blue}{Blue} and \textcolor{soft_red}{Red} dots represent the \textcolor{soft_blue}{preserved tokens $\mathbf{V}_P$} and \textcolor{soft_red}{edited tokens $\mathbf{V}_E$}. As illustrated, at $t=0.5T$, the overlap between \textcolor{soft_blue}{$\mathbf{V}_P$} and \textcolor{soft_red}{$\mathbf{V}_E$} becomes irreversible due to the contraction of the movement region.}
\label{fig:repaint_results}
\end{figure}

\begin{figure}[t]
    \centering
    \includegraphics[width=0.9\linewidth]{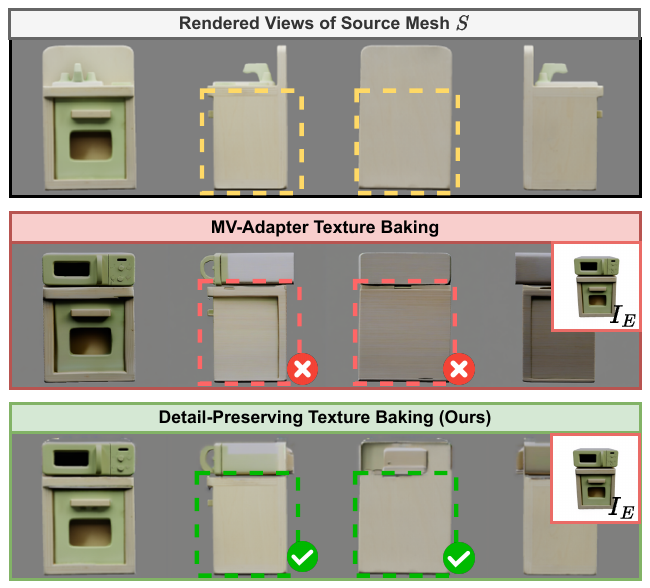}
    \caption{\textbf{Illustration of our proposed Detail-Preserving Texture Baking.} Relying solely on the standard MV-Adapter leads to visual discrepancies in the preserved regions (\textcolor{red!80}{highlighted in red box}). In contrast, our Detail-Preserving Texture Baking effectively mitigates these errors, maintaining the fidelity of the original unedited areas (\textcolor{green!60!black}{highlighted in green box}).}
    \label{fig:baking_method_demo}
\end{figure}

{
\begin{table*}[ht]
\small
\centering
\caption{\textbf{Quantitative comparison on Edit3D-Bench.} We evaluate preservation quality on unedited regions using Chamfer Distance (CD), PSNR, SSIM, and LPIPS. Additionally, we assess condition alignment using DINO-I and CLIP-T. The \textbf{best} and \underline{second best} results are highlighted in bold and underlined, respectively. We also provide the reconstruction performance of our backbone, \textcolor{gray}{TripoSG}, which serves as the upper bound for preservation quality.}
\label{tab:quantitative_comparison}
\begin{tabular}{lccccccccc}
\toprule
\multirow{2}{*}{Method} & \multirow{2}{*}{Time} & \multicolumn{5}{c}{Unedited Region Preservation} & \multicolumn{2}{c}{Condition Alignment} \\ \cmidrule(lr){3-7}\cmidrule(lr){8-9}
 & & CD $\downarrow$ & PSNR (M) $\uparrow$ & SSIM (M) $\uparrow$ & LPIPS (M) $\downarrow$ & FID $\downarrow$ & DINO-I $\uparrow$ & CLIP-T $\uparrow$ \\ 
\midrule
MVEdit~\cite{mvedit} & $\sim 160s$& 0.188 & 21.90 & 0.91 & 0.13 & 47.05 & 0.81 & 27.03 \\
Instant3DiT~\cite{barda2025instant3dit} & $\sim 20s$& 0.124 & 16.76 & 0.81 & 0.28 & 72.13 & 0.71 & 25.71  \\
Trellis~\cite{trellis}  &$\sim 600s$& \underline{0.014} & \underline{29.22}& \textbf{0.97} & \textbf{0.04} & 33.09 & \underline{0.91} & \underline{27.87} \\
VoxHammer~\cite{voxhammer}& $\sim 600s$ & 0.018 & 27.05 & \underline{0.95} & \underline{0.05} & \textbf{31.13} & 0.90 & \textbf{28.08} \\
\midrule
\textbf{Ours} & $\sim 200s$&\textbf{0.011} & \textbf{29.63} & \textbf{0.97} & \textbf{0.04} & \underline{32.63} & \textbf{0.92} & 27.75 \\
\textcolor{gray}{TripoSG (VAE Encode+Decode)~\cite{triposg}}& $\sim 100s$ & \textcolor{gray}{0.006} & \textcolor{gray}{31.88} & \textcolor{gray}{0.98} & \textcolor{gray}{0.02} & \textcolor{gray}{16.21} & - & - \\
\bottomrule
\end{tabular}
\end{table*}
}


\subsubsection{Drift-aware Token Pruning.}
Due to the spatial mobility of VecSet tokens, early-stage denoising may cause editable tokens to drift into regions that should remain unchanged, leading to geometric overlap that cannot be corrected in diffusion steps. A toy example of this scenario is shown in~\cref{fig:repaint_results}. Thus, we introduce \emph{Drift-aware Token Pruning}, which intervenes at a timestep $T_{\text{pruning}} \leq T_{\text{repaint}}$ to explicitly resolve conflicts between edited and preserved geometry. 

At the pruning timestep, we identify two complementary token subsets.  (i) $\mathbf{V}_{\text{cond}}$, contains tokens that are supported by the target image condition and thus should be retained for editing. (ii) $\mathbf{V}_{\text{conflict}}$, contains tokens that are structurally associated with the preserved region and therefore risk introducing geometric redundancy. Therefore, tokens in $\mathbf{V}_{\text{conflict}}$ should be removed to prevent geometric interference with the preserved region; however, those that are also supported by the image condition (\ie belonging to $\mathbf{V}_{\text{cond}}$) are essential for realizing the target edit and must be retained:
\begin{equation} \small
\label{eq:pruning}
    \mathbf{V}_{E}^{(T_{\text{pruning}})} \leftarrow
    \mathbf{V}_{E}^{(T_{\text{pruning}})} \setminus
    \left(\mathbf{V}_{\text{conflict}} \setminus \mathbf{V}_{\text{cond}}\right).
\end{equation}
The two sets are defined as:
\begingroup
\small
\begin{align}
    \mathbf{V}_{\text{cond}} &=
    \texttt{TokenSeeding}(\mathbf{V}_{\text{RP}}^{(T_{\text{pruning}})}, I_E, M_I), \\
    \mathbf{V}_{\text{conflict}} &=
    \texttt{TokenGating}(\mathbf{V}_{\text{RP}}^{(T_{\text{pruning}})}, \mathbf{V}_P).
\end{align}
\endgroup
This pruning strategy removes geometrically conflicting tokens while safeguarding tokens that are necessary to satisfy the edit, thereby maintaining both structural integrity and semantic fidelity.

Final pipeline of the VecSet-Edit process concludes by running the RePaint loop (from $T_{\text{repaint}}$ to $0$) with pruning at $T_{\text{pruning}}$ (detailed algorithm can be found in Algorithm 1 of Appendix D). 

\subsection{Detail-Preserving Texture Baking}
\label{app:texture}
Given our focus on editing meshes via single-image instructions, we prioritize methods designed for single-view conditioned texture synthesis~\cite{hunyuan3d20,mvpaint}. Specifically, we adopt MV-Adapter~\cite{mv_adapter} as our texture backbone.

\subsubsection{MV-Adapter}
The MV-Adapter~\cite{mv_adapter} module operates in two stages. The first stage, \textit{Image-Geometry-to-Multiview}, begins by rendering 6-view surface normals from the source mesh $\mathcal{S}$. These normals, combined with the condition image $I_E$, guide a fine-tuned text-to-image model (based on Stable Diffusion~\cite{sdxl} and IP-Adapter~\cite{ip-adapter}) to synthesize consistent multi-view RGB images. This generation process is governed by:
\begin{equation}
    \{I^\text{mv}_j\}_{j=1}^6 = \operatorname{MV-Adapter}(\{\mathcal{N}^{S}_j\}_{j=1}^6, I_E),
\end{equation}
where $\mathcal{N}^{S}_j$ denotes the rendered surface normal for view $j$. 

The second stage, \textit{Texture Projection}, projects these generated views onto the mesh surface. Utilizing a differentiable renderer, we perform gradient-based inverse rendering to optimize the UV texture map, ensuring the rendered appearance aligns with the generated multi-view images:
\begin{equation}
    \mathcal{S}^\text{textured} = \operatorname{TextureProjection}(\mathcal{S}, \{I^\text{mv}_j\}_{j=1}^6).
\end{equation}

However, applying this global baking strategy naively can be suboptimal. As illustrated in \cref{fig:baking_method_demo}, regenerating the texture for the entire mesh often degrades the high-frequency details of preserved regions that should ideally remain unchanged.

\begin{figure}[t]
    \centering
    \includegraphics[width=\linewidth]{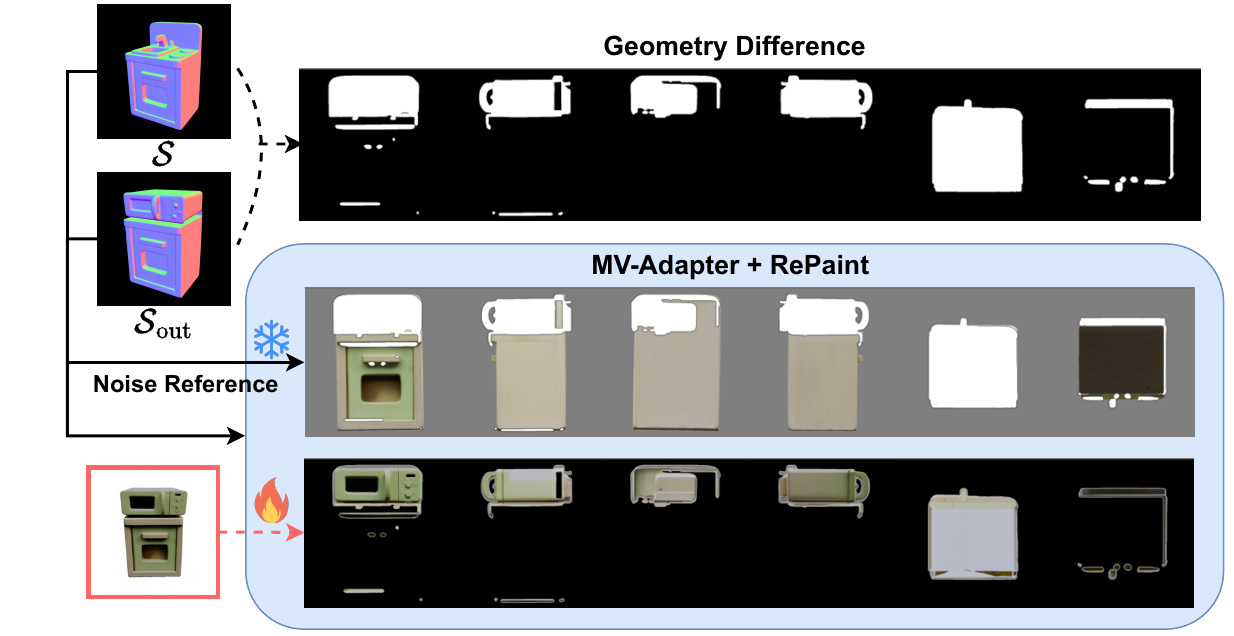}
    \caption{\textbf{Overview of the Detail-Preserving Texture Baking pipeline.} We compute geometric difference masks between the original and edited meshes to guide the MV-Adapter, ensuring that texture generation is restricted solely to the edited regions while preserving original details.}
    \label{fig:baking_method}
\end{figure}

\subsubsection{Geometry-aware Texture RePaint}
Leveraging the property that VecSet-Edit strictly preserves the geometry of the reference mesh $\mathcal{S}$ outside the edited region, we optimize the texturing process by exclusively updating areas with geometric changes.

As shown in \cref{fig:baking_method}, we quantify the geometric discrepancy between the source mesh $\mathcal{S}$ and the edited output $\mathcal{S}_\text{out}$ in the rendered view. This yields a set of difference masks $\{\mathbf{M}^\text{mv}_{j}\}_{j=1}^6$, defined as:
\begin{equation}
    \mathbf{M}^\text{mv}_{j} = \mathbb{I}\left( | \mathcal{N}^{S_{\text{out}}}_j - \mathcal{N}^{S}_j | > \tau_\text{texture} \right),
\end{equation}
where $\mathcal{N}_j$ represents the rendered normal map of view $j$, and $\tau_\text{texture}$ is the sensitivity threshold for detecting geometric shifts.

These masks serve as guidance for the MV-Adapter, restricting the generative process to the modified regions. This mechanism mirrors the logic of our VecSet RePaint strategy (\cref{eq:repaint}), but operates in the 2D pixel domain rather than the latent token space. Consequently, we effectively preserve the high-frequency texture details of the original mesh $\mathcal{S}$ while seamlessly propagating the semantic information from the condition image $I_E$ to the new geometry $\mathcal{S}_\text{out}$.

Formally, the multi-view in-painting process is defined as:
\begin{equation}
    \{I^\text{RePaint}_j\}_{j=1}^6 = \operatorname{MV-RePaint}(\{\mathcal{N}^{S_{\text{out}}}_j\}, \{\mathbf{M}^\text{mv}_{j}\}, I_E),
\end{equation}
\begin{equation}
    \mathcal{S}^\text{textured}_\text{out} = \operatorname{TextureProjection}(\mathcal{S}_\text{out}, \{I^\text{RePaint}_j\}_{j=1}^6).
\end{equation}

\begin{figure*}[ht]
\includegraphics[width=0.9\textwidth]{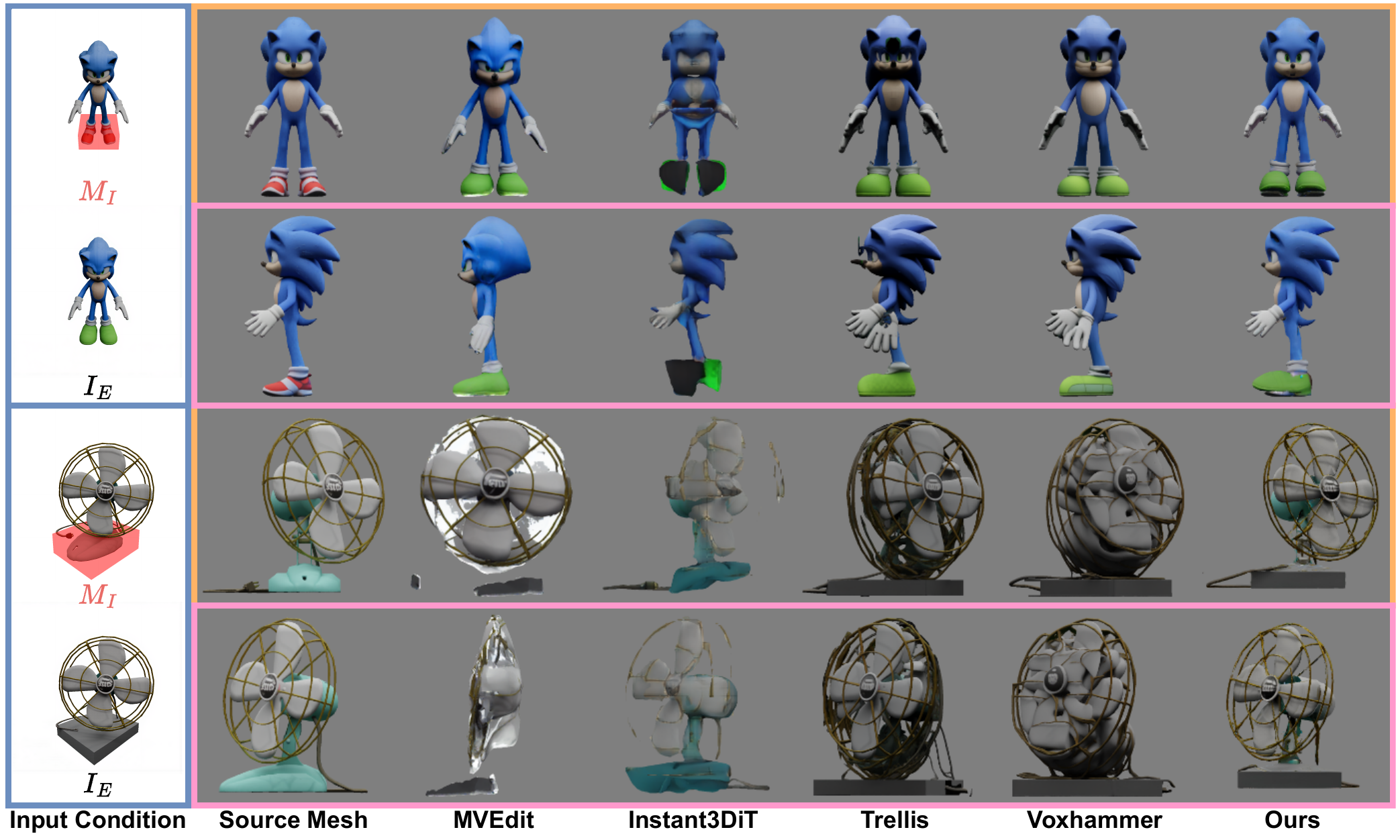}
\caption{\textbf{Qualitative comparison on Edit3D-Bench (The meshes are rendered from two views $0^\circ$ and $90^\circ$}). The results show that VecSet-Edit achieves superior preservation of source mesh details (such as hands and fans) while faithfully adhering to the target image condition.}
\label{fig:qualitative}
\end{figure*}

\section{Experiments}
We utilize TripoSG~\cite{triposg} as backbone. We set the RePaint starting timestep $T_\text{repaint}=0.7$, the pruning timestep $T_\text{pruning}=0.6$. The remaining settings and analysis can be found in Appendix D. We evaluate VecSet-Edit on \textbf{Edit3D-Bench~\cite{voxhammer}}, which is an established dataset for mesh editing containing 300 samples. Each sample provides a source mesh, a 3D bounding box, and an editing image. \emph{Notably, distinct from baselines that rely on the 3D bounding box for localization, our method uses the box only for evaluation.}

\subsection{Experimental Results}
 A core challenge in 3D editing is preventing unintended alterations. As presented in \cref{tab:quantitative_comparison}, by leveraging the VecSet-based backbone, our method achieves the lowest Chamfer Distance (CD)—21\% lower than the previous SOTA—along with superior image-based metrics (PSNR, SSIM, LPIPS). This indicates that VecSet-Edit effectively maintains the structural integrity and visual fidelity of unedited regions. In contrast, multi-view methods (e.g., MVEdit, Instant3DiT) often suffer from global distortions due to mesh misalignment, while voxel-based approaches (e.g., VoxHammer) exhibit higher CD errors attributable to limited grid resolution. \textbf{Beyond preservation quality}, VecSet-Edit achieves a $2\times$ wall-clock speedup compared to Trellis and VoxHammer, further validating the efficiency advantages of the VecSet backbone. Additionally, we utilize DINO-I and CLIP-T to evaluate visual and semantic consistency. Because our pipeline is designed for image-guided editing, we observe a significant advantage in DINO-I scores, indicating that our geometry faithfully captures the structural details of the input image. While our CLIP-T scores are slightly lower than other baselines, they remain competitive, reflecting our design priority: optimizing for strict visual (image) alignment rather than broad textual correspondence.

To further complement the evaluation, we conducted a user study with 12 3D practitioners. Participants evaluated 10 editing cases and selected the best result among baselines based on two criteria (120 total votes per criterion). The overwhelming preference of domain experts (\cref{tab:user_study}) confirms that our method yields superior localized editing quality. As evidenced in \cref{fig:qualitative}, \textbf{VecSet-Edit} visually outperforms existing methods. \textbf{Structurally}, the topological flexibility of our backbone allows us to excel at retaining intricate geometries, such as hands (rows 1, 2) and fans (rows 3, 4). Notably, we achieve robust preservation of unedited regions, surpassing the voxel-based SOTA, VoxHammer, without relying on explicit 3D bounding boxes. This visual success underscores the precision of our \emph{Token Selection} strategy and the effectiveness of \emph{Token Pruning} in eliminating geometric artifacts. \textbf{Texturally}, our method ensures high fidelity. By incorporating \emph{Detail-Preserving Texture Baking}, we successfully maintain original appearance details, preventing the texture blurring observed in other baselines.

\begin{table}[tb] 
  \centering 
  \caption{\textbf{User preference study on editing results.}}
  \label{tab:user_study} 
  \small
  \begin{tabular}{l c c c c}
    \hline
    \textbf{User Criterion} & \textbf{Ours} & VoxHammer & MVEdit & I3DiT \\
    \hline
    \textbf{Region Preservation} & \textbf{59.17\%} & 21.67\% & 10.83\% & 8.33\% \\
    \textbf{Condition Alignment} & \textbf{58.33\%} & 26.67\% & 14.17\% & 0.83\% \\
    \hline
  \end{tabular}
\end{table}



\subsection{Ablation Study} We quantitatively dissect the contribution of each component in \cref{tab:ablation_quantitative}. Initially, the integration of \textbf{Token Seeding} and \textbf{Token Gating} yields a marked improvement in preservation metrics, demonstrating that 2D image conditions alone suffice for precise region localization. Subsequently, the application of \textbf{Token Pruning} significantly mitigates geometric errors (reflected in lower CD scores) by actively filtering out redundant tokens that drift into the preserved regions. Finally, our \textbf{Detail-Preserving Texture Baking} module serves as the final refinement step, without this step we can observe the decline in unedited region preservation.

\begin{table}[t]
    \caption{\textbf{Quantitative results of ablation study on Vecset Edit.}} 
        \label{tab:ablation_quantitative}
        \small 
        \begin{tabular}{lccccc}
            \toprule
            &CD&PSNR&LPIPS&DINO-I&CLIP-T \\
            \midrule
            RePaint&0.024 &24.35&0.07&0.88&27.25 \\
            \midrule
            + Token Seeding & 0.006&26.32&0.04&0.79&26.19 \\
            + Token Gating & 0.011&24.71&0.06&0.85&27.36 \\
            + Token Pruning &0.011&25.17&0.06&0.89&27.68 \\
            + Detail-Preserving (Full)&0.011&29.63&0.04&0.92&27.75 \\
            \bottomrule
        \end{tabular}
\end{table}
\section{Conclusion}
We present \textbf{VecSet-Edit}, the first training-free framework for localized mesh editing within VecSet-based LRM latent spaces. By uncovering the \textbf{VecSet Geometry Property}, we exploit the spatial locality of unordered tokens for precise manipulation. We propose \textbf{Mask-guided Token Seeding} and \textbf{Attention-aligned Token Gating} to localize edits using only 2D supervision, coupled with \textbf{Drift-aware Token Pruning} to ensure geometric consistency during diffusion. Finally, our \textbf{Detail-preserving Texture Baking} maintains details in unedited regions. Evaluations on Edit3D-Bench demonstrate that VecSet-Edit significantly outperforms existing approaches in preservation and alignment, bridging the gap between generative 3D models and controllable production workflows.
\section*{Acknowledgements}
This work is partially supported by the National Science and Technology Council, Taiwan, under Grant: NSTC-112-2221-E-A49-059-MY3 and NSTC-112-2221-E-A49-094-MY3.


\clearpage
\appendix
\makeatletter
\newcommand{\tabcaption}{\def\@captype{table}\caption}
\makeatother
\begin{figure*}[p] 
    \centering
    
    \includegraphics[width=0.88\linewidth]{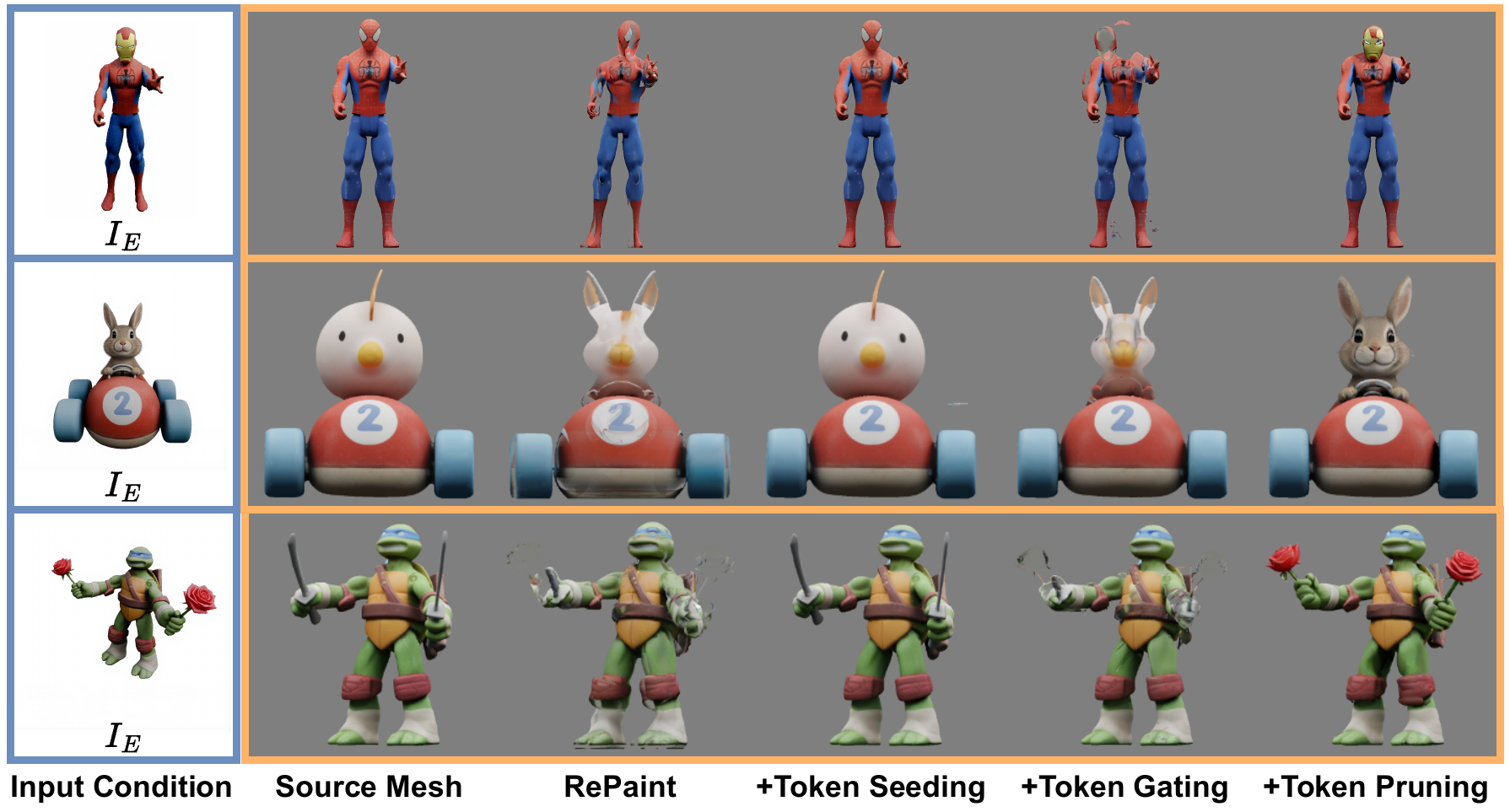}
   \caption{\textbf{Visual ablation of VecSet-Edit.} We demonstrate the necessity of modules: \textit{Token Seeding} localizes the edit to prevent the global distortion seen in naive RePaint; \textit{Token Gating} expands the selection to ensure coverage of the target region; and \textit{Token Pruning} removes outlier tokens and geometric artifacts.}
    \label{fig:ablation visualization}
    
    
    \begin{minipage}[t]{0.48\linewidth}
        \centering
        
        \includegraphics[width=\linewidth]{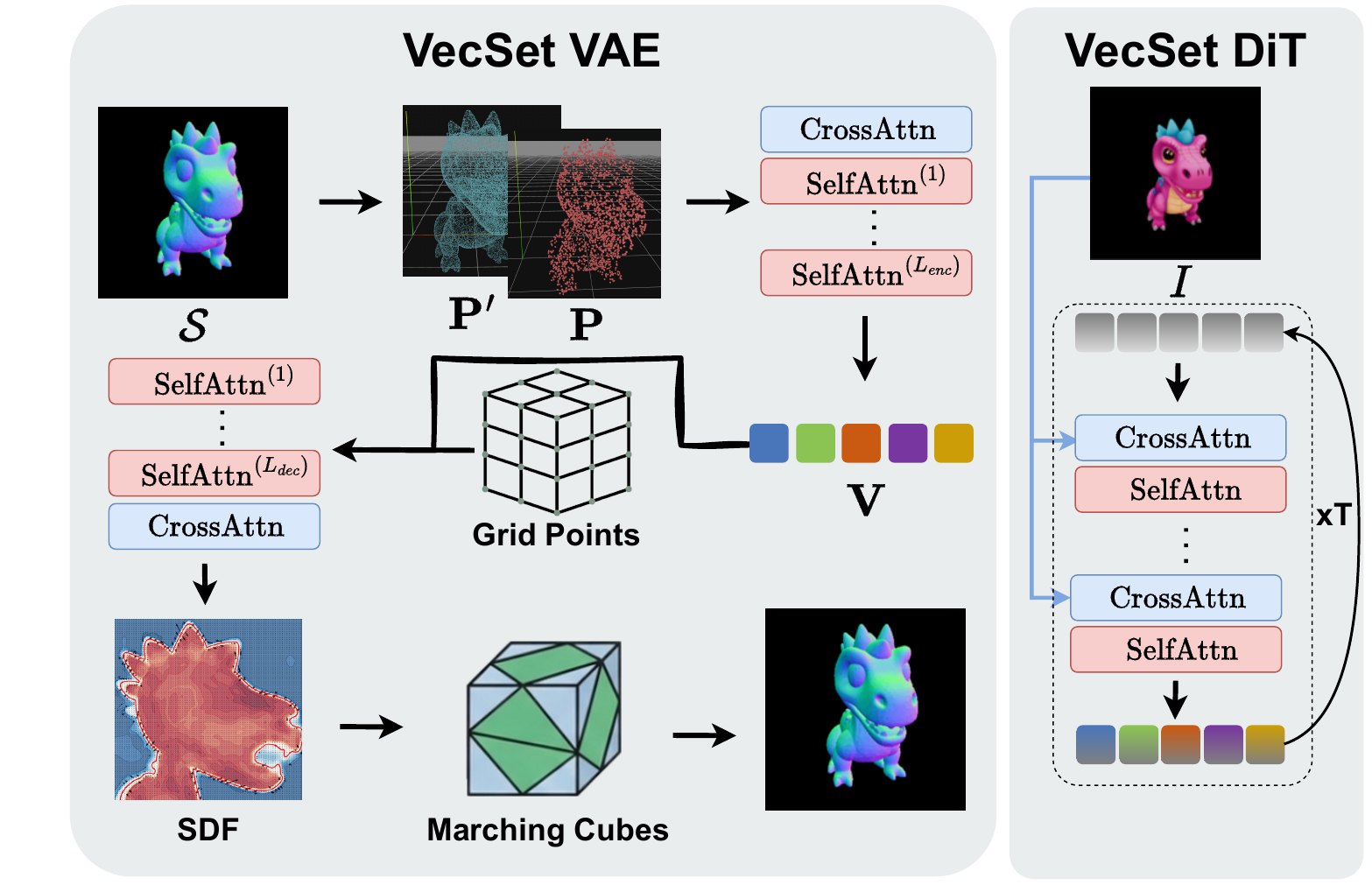}
        \vspace{-6mm}
        \caption{\textbf{Illustration of VecSet VAE and VecSet DiT.}}
        \label{fig:vecset_diffusion}
        
        
        \includegraphics[width=\linewidth]{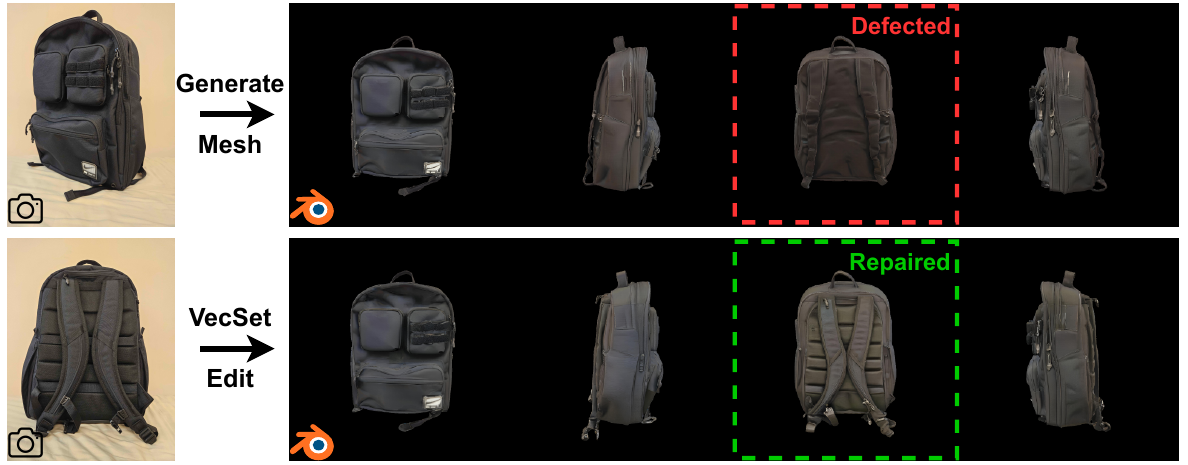}
        \caption{\textbf{Illustration of using VecSet-Edit to calibrate mesh.} While the generated mesh via TripoSG yields defective geometry due to single-view limitations (row 1), VecSet-Edit utilizes additional views to refine these defects, ensuring global consistency while preserving the original mesh details.}
        \label{fig:demo_iedit3d}
    \end{minipage}
    \hfill 
    \begin{minipage}[t]{0.48\linewidth}
        \centering
        

        \includegraphics[width=\linewidth]{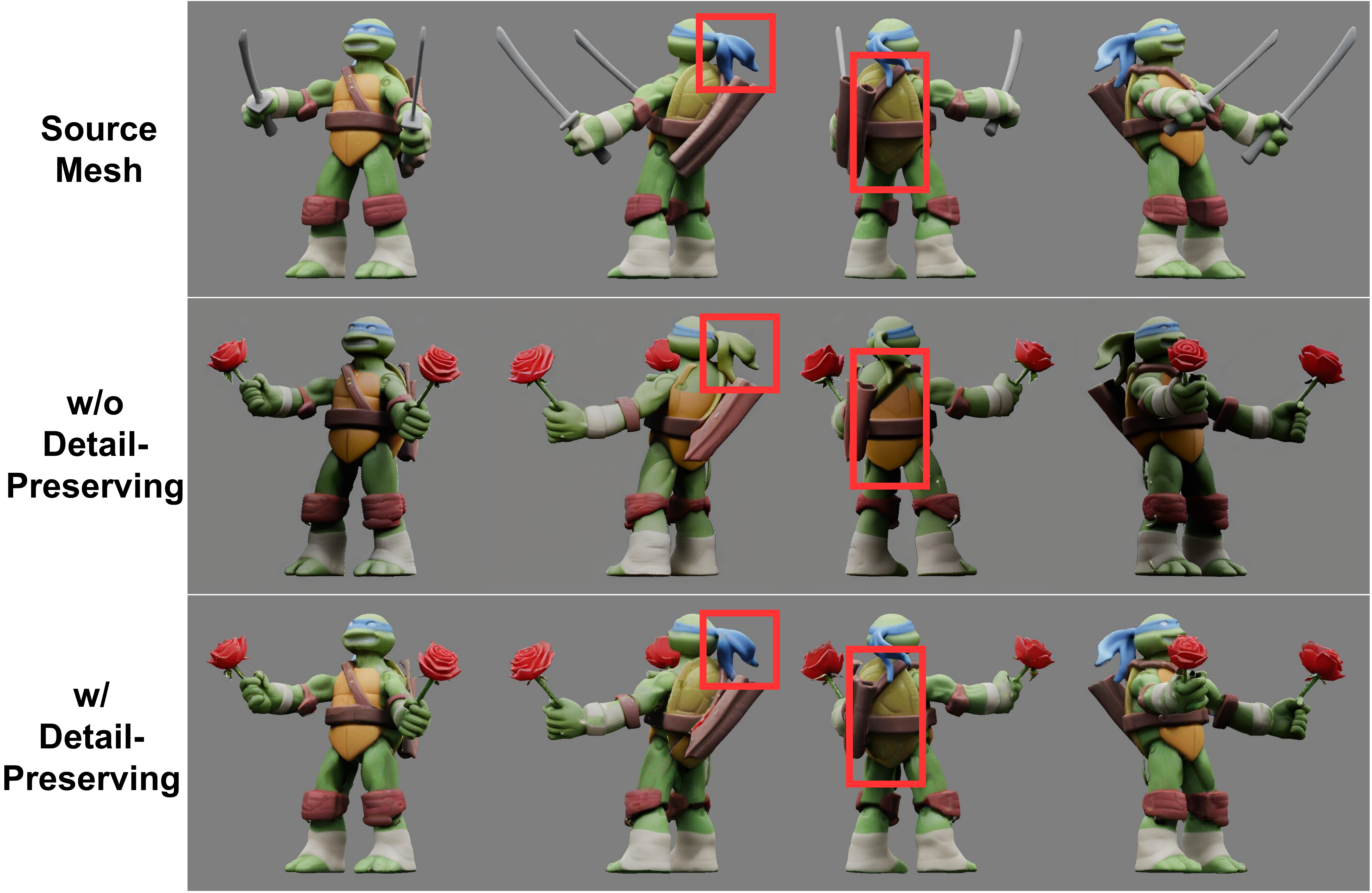}
        \caption{\textbf{Visual ablation of Detail-Preserving Texture Baking.} Our proposed baking strategy faithfully maintains the color fidelity of the edited mesh while preserving high-frequency details, even in views not visible in the editing condition $I_E$. \textcolor{red}{Key differences are highlighted in red boxes.}} \label{fig:texture_ablation}
    \end{minipage}
\end{figure*}
\begin{figure*}
    \includegraphics[width=\linewidth]{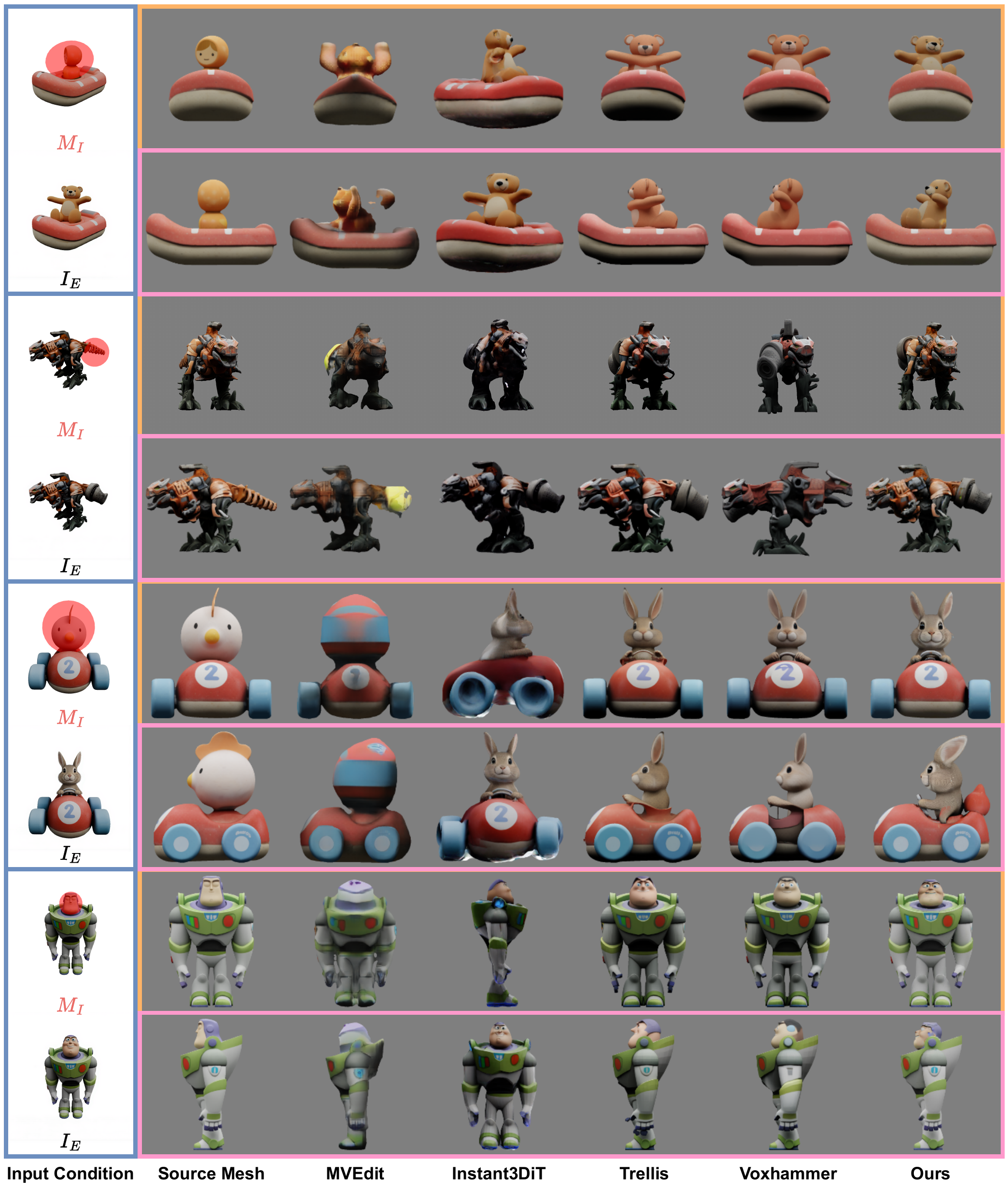}
    \caption{\textbf{More qualitative comparison on Edit3D-Bench.} Our proposed VecSet-Edit shows superior performance across different input scenarios.}
    \label{fig:qualitative_plus}
    
\end{figure*}







\clearpage
\bibliographystyle{ACM-Reference-Format}
\bibliography{sample-base}

\clearpage
\begin{appendices}
    \section{Details of VecSet-based LRM}
\label{app:vecset}
The illustration of overall VecSet Encoding and Diffusion process can be found in \cref{fig:vecset_diffusion}, other details can be found below.
\subsection{VecSet VAE}
\label{sec:vae}
To encode a mesh $\mathcal{S}$, we first sample a point cloud $\mathbf{P} \in \mathbb{R}^{N_{p}\times 6}$ from the mesh surface, containing both position and normal information. A learnable subset of tokens, initialized as $\mathbf{P}' \in \mathbb{R}^{N_{p'}\times 6}$, serves as the initial latent queries (typically $N_p=50,000$ and $N_{p'}=2048$). These surface points are integrated into the latent queries via cross-attention to produce the VecSet tokens $\mathbf{V}$, which encode the geometric information of the input mesh:
\begin{align}
\label{eq:encode}
\mathbf{V}' &= \operatorname{CrossAttn}(\operatorname{PosEmb}(\mathbf{P}'), \operatorname{PosEmb}(\mathbf{P})), \\
\mathbf{V} &= \operatorname{SelfAttn}^{(i)}(\mathbf{V}'), \quad i \in \{1, \dots, L_{\text{enc}}\},
\end{align}
where $\operatorname{CrossAttn}(Q, K)$ denotes a cross-attention layer where queries attend to keys, and $\operatorname{SelfAttn}^{(i)}$ represents the $i$-th self-attention layer.

Once the VecSet representation $\mathbf{V}$ is obtained, we can decode it into a Signed Distance Field (SDF). For an arbitrary query position $x \in \mathbb{R}^3$, the predicted signed distance $d$ is computed as:
\begin{align}
\widetilde{\mathbf{V}} &= \operatorname{SelfAttn}^{(i)}\left(\operatorname{Linear}(\mathbf{V})\right), \quad i \in \{1, \dots, L_{\text{dec}}\}, \\
d(x) &= \operatorname{CrossAttn}\left(\operatorname{PosEmb}(x), \widetilde{\mathbf{V}}\right).
\end{align}
Finally, the explicit mesh $\mathcal{S}'$ is extracted by applying Marching Cubes~\cite{marching_cubes} on the predicted SDF at a specified resolution. 
In the following discussion, we abbreviate the VecSet encoding process as $\operatorname{Encode}$ and the decoding pipeline (including SDF prediction and Marching Cubes) as $\operatorname{Decode}$, \ie $\mathcal{S}_{\text{R}} = \operatorname{Decode}(\operatorname{Encode}(\mathcal{S})).$

\section{Details of Token Selection}\label{appdx:sharpness}
\subsection{Attentive-Layer Selection}
Inspired by prior text-to-image editing works~\cite{prompt_to_prompt,tf_ti2i,tf-gph,diffedit,sdedit,multi_turn,kulikov2025flowedit,cao2023masactrl}, which demonstrate that cross-attention maps provide effective zero-shot cues for semantic localization, we adapt this principle to the 3D VecSet DiT framework to identify geometry-relevant tokens from a 2D edit mask.

Recall from \cref{eq:cross_attn} that VecSet tokens $\mathbf{V}$ attend to image features $h_I$ through cross-attention. We denote the resulting cross-attention map as
\begin{equation}
    \mathbf{A}^{(l,t)}_{\text{cross}} \in \mathbb{R}^{N_p \times (HW)},
\end{equation}
where $N_p$ is the number of VecSet tokens, and $(l,t)$ index the DiT block and diffusion timestep, respectively. Each entry $\mathbf{A}^{(l,t)}_{\text{cross}}[i,j]$ measures the attention weight from the $i$-th 3D token to the $j$-th image pixel.

To quantify the relevance of a VecSet token to the user-specified edit region, we aggregate its attention over the masked pixels:
\begin{equation}
    a^{(l,t)}_M(i) = \sum_{j=1}^{HW} \mathbf{A}^{(l,t)}_{\text{cross}}[i,j] \cdot \bar{M_I}[j],
\end{equation}
where $a^{(l,t)}_M(i)$ is a scalar that reflects how strongly the $i$-th token aligns with the masked region of $I_S$ and $\bar{M}_I \in \{0, 1\}^{HW}$ is flatten from $M_I$.\footnote{Attention values are averaged across heads in practice.}

A naive strategy is to average $a^{(l,t)}_M(i)$ across all blocks and timesteps to obtain a global relevance score. However, we observe that semantic alignment varies significantly across DiT blocks, consistent with findings in prior diffusion analyses~\cite{multi_turn}. Some layers exhibit sharp, condition-aware attention, while others produce diffuse or uninformative patterns.

To identify informative layers, we analyze the sharpness of their cross-attention distributions. For each DiT block $l$, we compute the Kullback--Leibler (KL) divergence between token-wise attention and its marginal distribution:
\begin{equation}
    \mathcal{D}^{(l)} = \sum_t\sum_{i=1}^{N_p} \operatorname{KL}\!\left(
    \mathbf{A}^{(l,t)}_{\text{cross}}[i,:] \,\middle\|\, \bar{\mathbf{a}}^{(l,t)}_{\text{M}}
    \right),
\end{equation}
where $\bar{\mathbf{a}}^{(l)}_{\text{cross}} \in \mathbb{R}^{HW}$ is the marginal attention defined by
\begin{equation}
    \bar{\mathbf{a}}^{(l,t)}_{\text{cross}}[j] = \sum_{i=1}^{N_p} \mathbf{A}^{(l,t)}_{\text{cross}}[i,j].
\end{equation}
Higher $\mathcal{D}^{(l)}$ indicates a sharper and more semantically aligned attention pattern (see \cref{fig:kl}). We therefore select the top-$K$ most informative layers:
\begin{equation}
    L_{\text{attn}} = \operatorname{TopK}_l \big( \mathcal{D}^{(l)} \big).
\end{equation}

\subsection{Adaptive Threshold}

Using only these layers, we compute the refined relevance score by averaging across selected layers and timesteps:
\begin{equation}
    \tilde{a}_M(i) = \mathbb{E}_{l \in L_{\text{attn}},\, t}\!\left[ a^{(l,t)}_M(i) \right].
\end{equation}
Empirically, $\tilde{a}_M$ exhibits a long-tailed distribution, where only a small subset of tokens strongly aligns with the edit region. We therefore define an adaptive threshold based on the top-$\alpha\%$ scores:
\begin{equation}
    \tau_I(\tilde{a}_M, \alpha_I) =
    \frac{\alpha_I}{100} \cdot
    \operatorname{mean}\!\left(
    \operatorname{Top}_{10\%}(\tilde{a}_M)
    \right),
    \label{eq:threshold}
\end{equation}
similarily the computation of $\tau_A$ is also following \cref{eq:threshold} and control by $\alpha_A$.

\section{Detail-Preserving Texture Baking}
\label{app:texture}
Texture baking is a fundamental step in automated mesh generation pipelines~\cite{triposg,hunyuan3d_omni,hunyuan3d20,mv_adapter,mvpaint,paint3d,dreammat,texture}. Given our focus on editing meshes via single-image instructions, we prioritize methods designed for single-view conditioned texture synthesis~\cite{triposg,hunyuan3d20,mv_adapter,mvpaint}. Specifically, we adopt MV-Adapter~\cite{mv_adapter} as our texture backbone.

\subsection{MV-Adapter}
The MV-Adapter~\cite{mv_adapter} module operates in two stages. The first stage, \textit{Image-Geometry-to-Multiview}, begins by rendering 6-view surface normals from the source mesh $\mathcal{S}$. These normals, combined with the condition image $I_E$, guide a fine-tuned text-to-image model (based on Stable Diffusion~\cite{sdxl} and IP-Adapter~\cite{ip-adapter}) to synthesize consistent multi-view RGB images. This generation process is governed by:
\begin{equation}
    \{I^\text{mv}_j\}_{j=1}^6 = \operatorname{MV-Adapter}(\{\mathcal{N}^{S}_j\}_{j=1}^6, I_E),
\end{equation}
where $\mathcal{N}^{S}_j$ denotes the rendered surface normal for view $j$. The six views consist of four horizontal azimuths $\{0^\circ, 90^\circ, 180^\circ, 270^\circ\}$ and two vertical elevations at $\{90^\circ, 270^\circ\}$.

The second stage, \textit{Texture Projection}, projects these generated views onto the mesh surface. Utilizing a differentiable renderer, we perform gradient-based inverse rendering to optimize the UV texture map, ensuring the rendered appearance aligns with the generated multi-view images:
\begin{equation}
    \mathcal{S}^\text{textured} = \operatorname{TextureProjection}(\mathcal{S}, \{I^\text{mv}_j\}_{j=1}^6).
\end{equation}

However, applying this global baking strategy naively can be suboptimal. As illustrated in \cref{fig:baking_method_demo}, regenerating the texture for the entire mesh often degrades the high-frequency details of preserved regions that should ideally remain unchanged.

\begin{figure}[t]
    \centering
    \includegraphics[width=\linewidth]{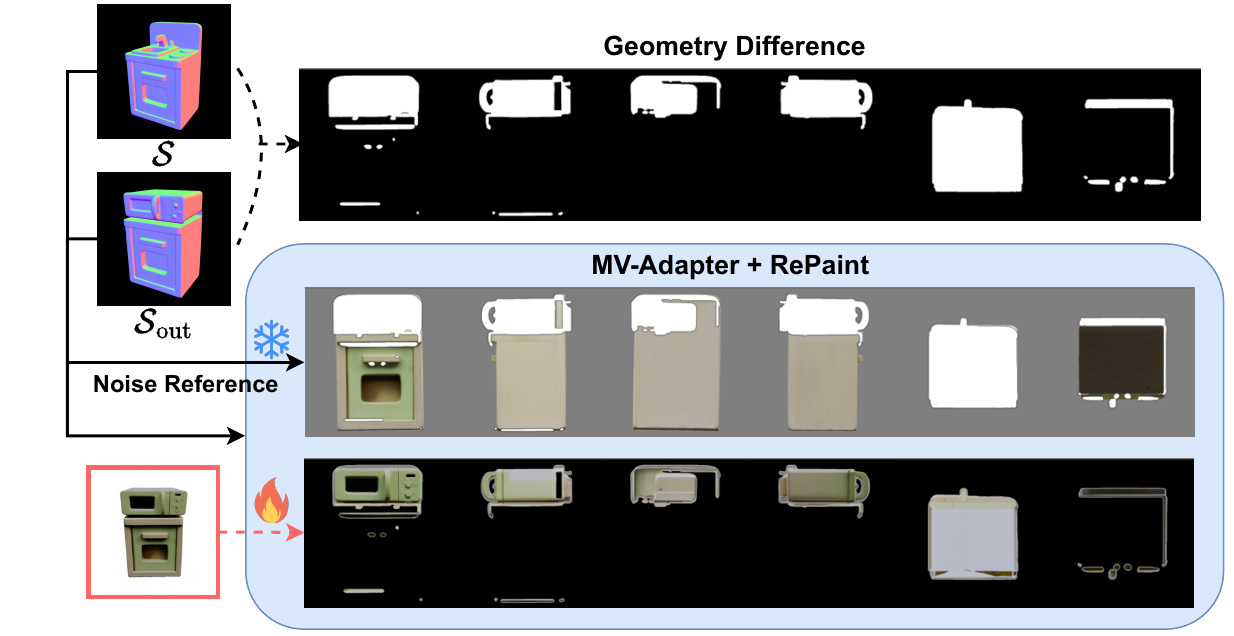}
    \caption{\textbf{Overview of the Detail-Preserving Texture Baking pipeline.} We compute geometric difference masks between the original and edited meshes to guide the MV-Adapter, ensuring that texture generation is restricted solely to the edited regions while preserving original details.}
    \label{fig:baking_method}
\end{figure}

\subsection{Geometry-aware Texture RePaint}
Leveraging the property that VecSet-Edit strictly preserves the geometry of the reference mesh $\mathcal{S}$ outside the edited region, we optimize the texturing process by exclusively updating areas with geometric changes.

As shown in \cref{fig:baking_method}, we first quantify the geometric discrepancy between the source mesh $\mathcal{S}$ and the edited output $\mathcal{S}_\text{out}$ in the rendered view space. This yields a set of difference masks $\{\mathbf{M}^\text{mv}_{j}\}_{j=1}^6$, defined as:
\begin{equation}
    \mathbf{M}^\text{mv}_{j} = \mathbb{I}\left( | \mathcal{N}^{S_{\text{out}}}_j - \mathcal{N}^{S}_j | > \tau_\text{texture} \right),
\end{equation}
where $\mathcal{N}_j$ represents the rendered normal map of view $j$, and $\tau_\text{texture}$ is the sensitivity threshold for detecting geometric shifts.

These masks serve as spatial guidance for the MV-Adapter, restricting the generative process to the modified regions. This mechanism mirrors the logic of our VecSet RePaint strategy (\cref{eq:repaint}), but operates in the 2D pixel domain rather than the latent token space. Consequently, we effectively preserve the high-frequency texture details of the original mesh $\mathcal{S}$ while seamlessly propagating the semantic information from the condition image $I_E$ to the new geometry $\mathcal{S}_\text{out}$.

Formally, the multi-view in-painting process is defined as:
\begin{equation}
    \{I^\text{RePaint}_j\}_{j=1}^6 = \operatorname{MV-RePaint}(\{\mathcal{N}^{S_{\text{out}}}_j\}, \{\mathbf{M}^\text{mv}_{j}\}, I_E),
\end{equation}
followed by the final texture projection:
\begin{equation}
    \mathcal{S}^\text{textured}_\text{out} = \operatorname{TextureProjection}(\mathcal{S}_\text{out}, \{I^\text{RePaint}_j\}_{j=1}^6).
\end{equation}

\section{VecSet-Edit Settings}\label{app:vecset_setting} 
\begin{figure*}
    \centering
    \includegraphics[width=\linewidth]{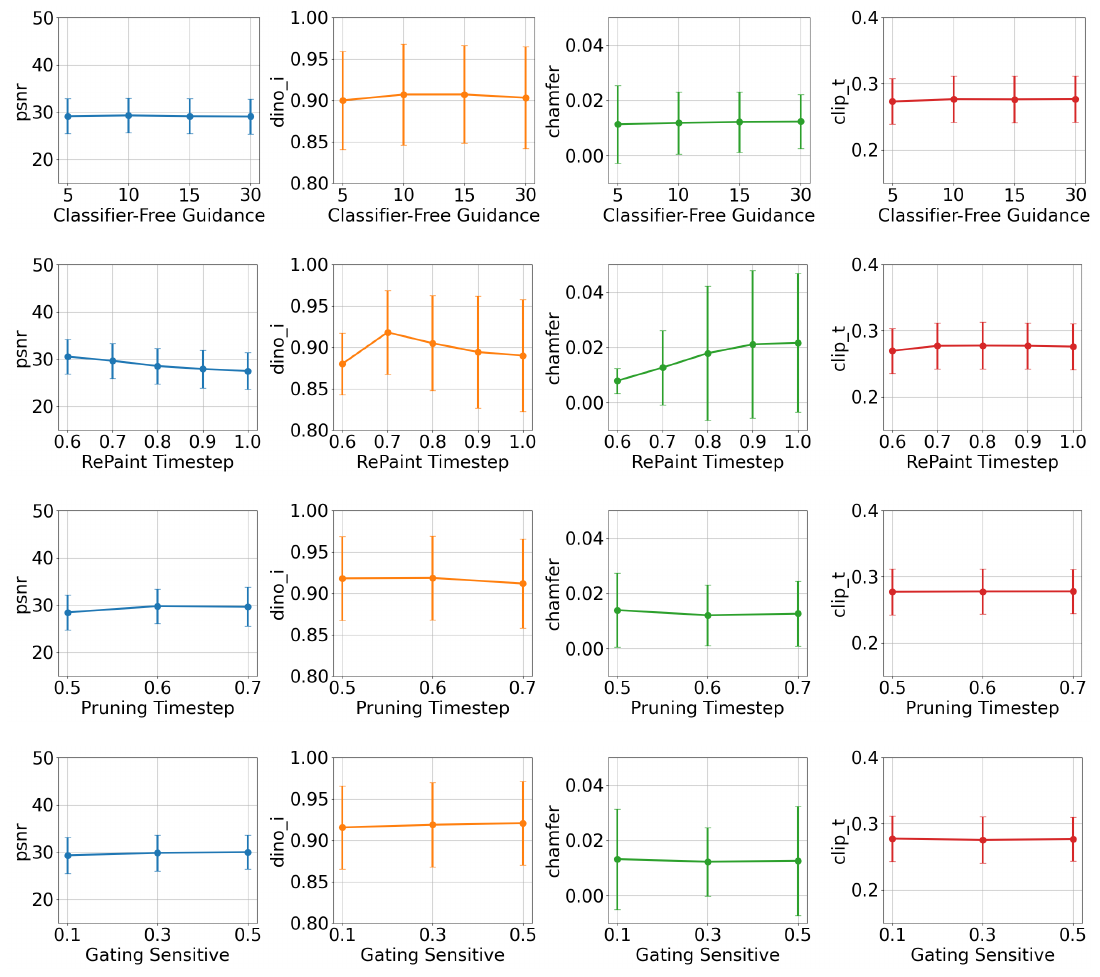}
    \caption{\textbf{Sensitive test of VecSet-Edit hyperparameters.}}
    \label{fig:sensitive_test}
\end{figure*}

\begin{algorithm}[t]
\caption{VecSet-Edit}
\label{alg:vecset_edit}
\SetKwInOut{Input}{Input}
\SetKwInOut{Output}{Output}
\SetKwFunction{Gather}{Gather}
\SetKwFunction{ImageSel}{TokenSeeding}
\SetKwFunction{SelfSel}{TokenGating}

\Input{Reference VecSet $\mathbf{V}$, source image $I_S$, edit image $I_E$, mask $M_I$, repaint start $T_{\text{repaint}}$, pruning time $T_{\text{pruning}}$}
\Output{Edited VecSet $\mathbf{V}_{\text{out}}$}
$\mathbf{V}_I \leftarrow \ImageSel(\mathbf{V}, I_S, M_I)$\;
$\mathbf{V}_E \leftarrow \SelfSel(\mathbf{V}, \mathbf{V}_I)$\;
Let $\mathbb{I}_E$ be indices such that $\mathbf{V}_E=\Gather(\mathbf{V},\mathbb{I}_E)$\;
\BlankLine
Sample $\epsilon_P \sim \mathcal{N}(0,\mathbf{I})$\;
$\mathbf{V}_{\text{RP}}^{(T_{\text{repaint}})} \leftarrow (1-T_{\text{repaint}})\,\epsilon_P + T_{\text{repaint}}\,\mathbf{V}$\;
$\mathbf{V}_E^{(T_{\text{repaint}})} \leftarrow \Gather(\mathbf{V}_{\text{RP}}^{(T_{\text{repaint}})}, \mathbb{I}_E)$\;
$\mathbf{V}_P^{(T_{\text{repaint}})} \leftarrow \mathbf{V}_{RP}^{(T_{\text{repaint}})} \setminus V_E^{(T_{\text{repaint}})}$\;
\BlankLine
\For{$t = T_{\text{repaint}},\, T_{\text{repaint}}-\Delta t,\, \dots,\, \Delta t$}{
    $v_{\text{pred}} \leftarrow u_{\theta}(\mathbf{V}_{\text{RP}}^{(t)}, h_I, t)$ \tcp*{Full context prediction}
    $\mathbf{V}_E^{(t-\Delta t)} \leftarrow \mathbf{V}_E^{(t)} - \Gather(v_{\text{pred}}, \mathbb{I}_E)\cdot \Delta t$\;
    $\mathbf{V}_P^{(t-\Delta t)} \leftarrow (1-(t-\Delta t))\,\epsilon_P + (t-\Delta t)\,\mathbf{V}_P$\;

    \If{$t = T_{\text{pruning}}$}{
        $\mathbf{V}_{\text{cond}} \leftarrow \ImageSel(\mathbf{V}_{\text{RP}}^{(t-\Delta t)}, I_E, M_I)$\;
        $\mathbf{V}_{\text{conflict}} \leftarrow \SelfSel(\mathbf{V}_{\text{RP}}^{(t-\Delta t)}, \mathbf{V}_P)$\;
        $\mathbf{V}_E^{(t-\Delta t)} \leftarrow \mathbf{V}_E^{(t-\Delta t)} \setminus \left(\mathbf{V}_{\text{conflict}} \setminus \mathbf{V}_{\text{cond}}\right)$\tcp*{Pruning}
    }

    $\mathbf{V}_{\text{RP}}^{(t-\Delta t)} \leftarrow \mathbf{V}_E^{(t-\Delta t)} \oplus \mathbf{V}_P^{(t-\Delta t)}$\;
}
\Return{$\mathbf{V}_{\text{out}} \leftarrow \mathbf{V}_{\text{RP}}^{(0)}$}\;
\end{algorithm}
All experiments were conducted on NVIDIA HGX H100 4-Way. The VRAM requirement ranges from $22$ to $30$ GB depending on the input complexity. We summarize the default hyperparameter configurations of VecSet-Edit below. For the hyperparameter sensitivity analysis, we varied one specific parameter while keeping the others fixed at their default values, the evaluation results can be found in \cref{fig:sensitive_test}. \footnote{Sensitivity experiments were conducted on a randomly selected $50\%$ subset of the Edit3D-Bench to reduce computational overhead}

\subsection{LRM Backbone}
\noindent\textbf{Classifier-free Guidance Scale ($r=10$).} This parameter controls the alignment of the generated mesh with the condition image. Empirically, we observed that the mesh geometry remains relatively robust to variations in this parameter.

\noindent\textbf{Number of Inference Steps ($n_\text{step}=50$).} We adopt the default configuration from TripoSG~\cite{triposg}. While increasing this value typically improves generation quality, it also linearly increases inference latency.

\begin{figure}[t]
    \centering
    \includegraphics[width=\linewidth]{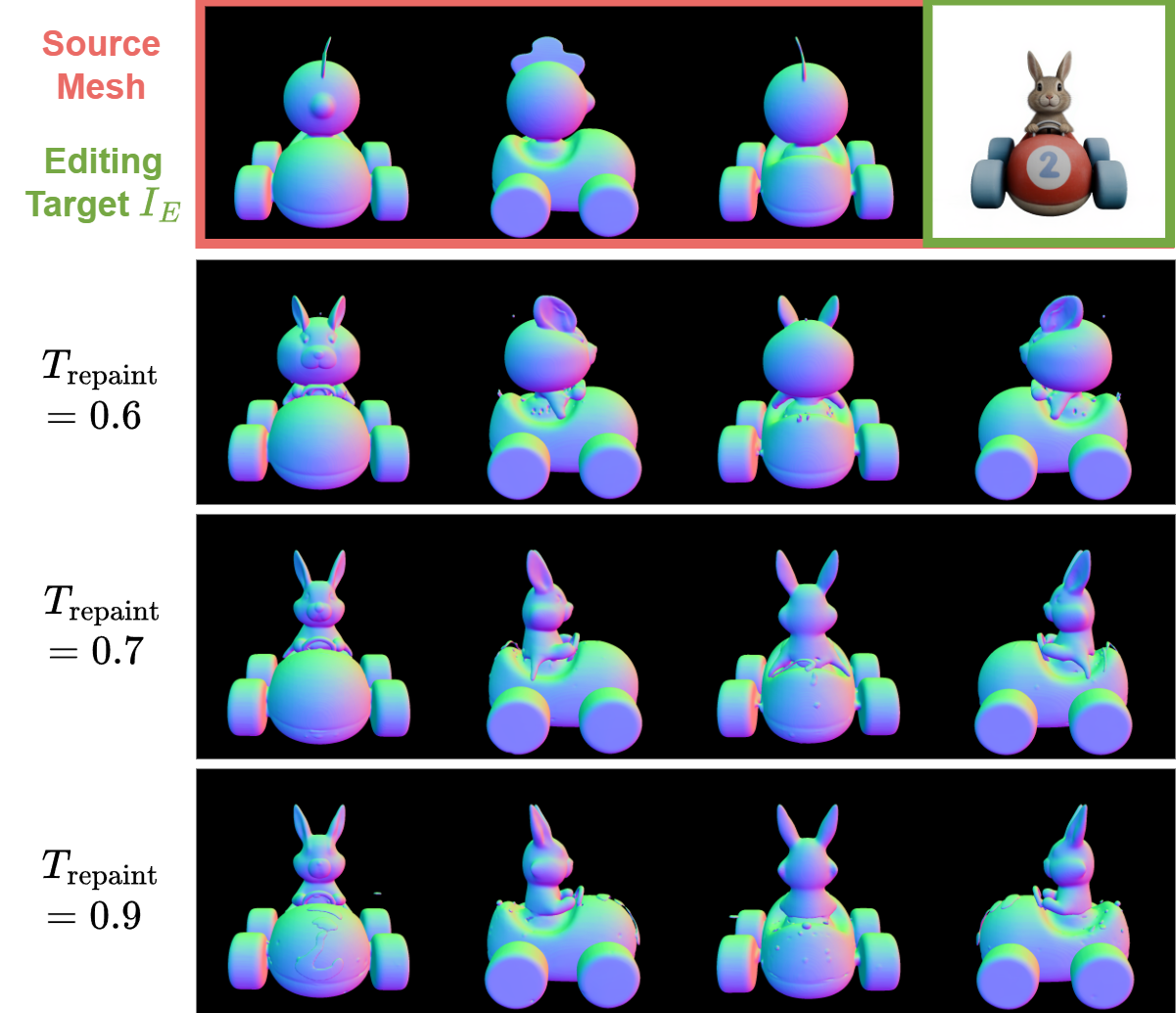}
    \caption{\textbf{Ablation study of the Mesh Editing hyperparameter \(T_\text{repaint}\).} Lowering \(T_\text{repaint}\) biases the output heavily toward the source mesh, resulting in an edit that fails to match the condition image. Conversely, an excessively high \(T_\text{repaint}\) leads to a general degradation in editing quality (e.g., loss of sharp facial details).}
    \label{fig:ablation_repaint}
\end{figure}

\begin{figure}[t]
    \centering
    \includegraphics[width=\linewidth]{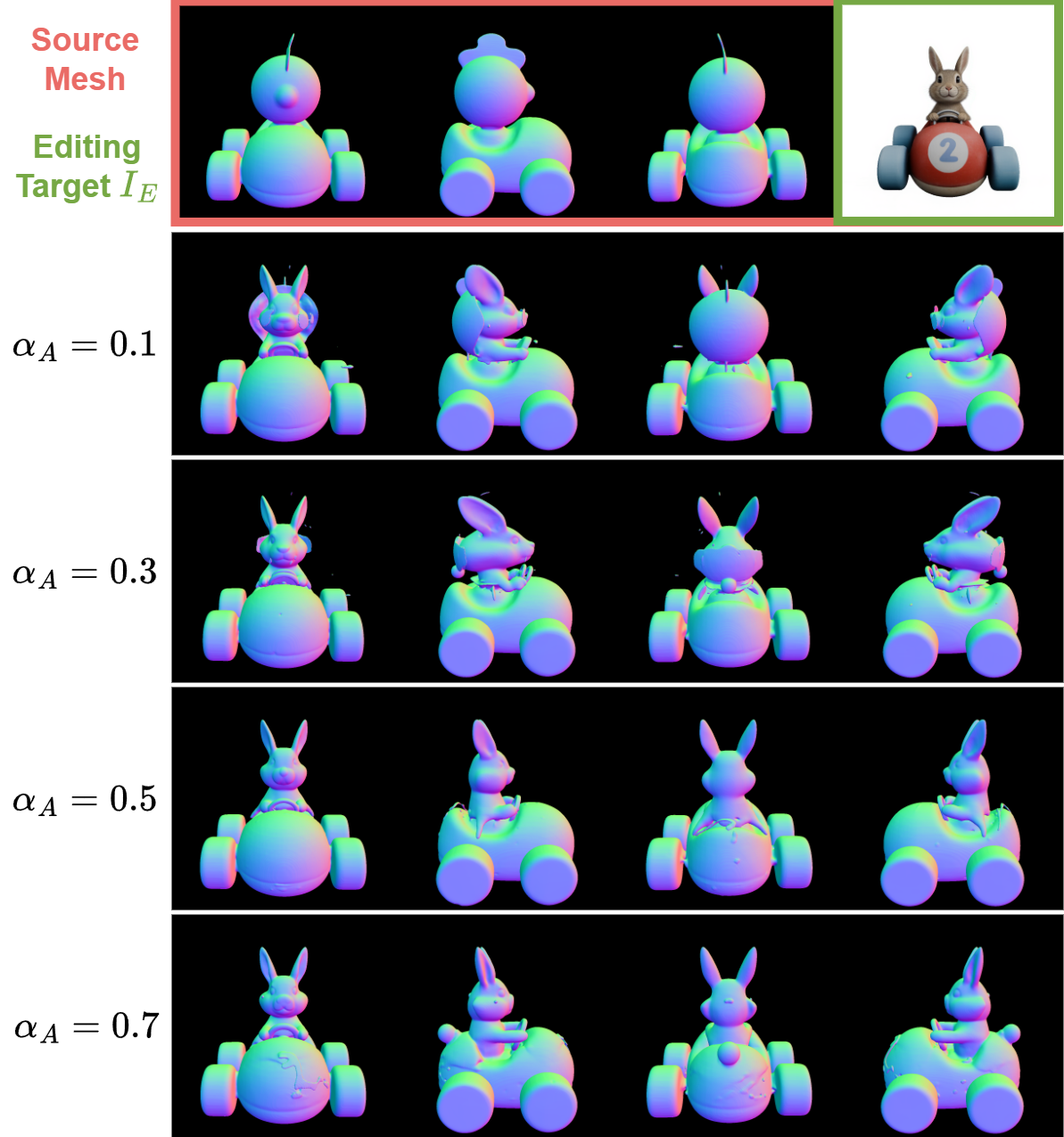}
    \caption{\textbf{Ablation study of the Token Gating hyperparameter \(\alpha_A\).} Lowering \(\alpha_A\) causes the edited mesh to retain an excess of preserved tokens, which ultimately leads to incomplete edits. Conversely, an excessively high \(\alpha_A\) leads to unintended distortion in regions that should remain unchanged (e.g., the car).}
    \label{fig:ablation_gating}
\end{figure}

\subsection{Editing Process}
\noindent\textbf{RePaint Timestep ($T_\text{repaint}=0.7$).} This parameter governs the editing strength. A lower $T_\text{repaint}$ constrains the output too strictly to the source mesh, limiting editability. Conversely, an excessively high $T_\text{repaint}$ removes the geometric prior, leading to incoherence between the edited and unedited regions. 
Crucially, this setting is \textit{intention-dependent}: for structural changes (e.g., removing an object), a higher $T_\text{repaint}$ is required; for tasks requiring strict coherence (e.g., local detail refinement), a lower value is preferred. The qualitative example can be found in \cref{fig:ablation_repaint}

\noindent\textbf{Pruning Timestep ($T_\text{pruning}=0.6$).} This determines when to filter out outlier tokens. Pruning too early (at high noise levels) compromises geometric quality, as valid detail tokens may be misclassified as outliers. Pruning too late leaves insufficient denoising steps for the VecSet to realign the remaining tokens, resulting in disconnected geometry.

\noindent\textbf{Seeding Sensitive ($\alpha_\text{I}=0.7$).} The seeding sensitive controls the precision of the initial token selection. We keep this value high to ensure high recall (i.e., accurately locating all potential image-related tokens). Extensive ablation was not performed on this parameter, as the final selection quality is primarily dominated by the subsequent gating step.

\noindent\textbf{Gating Sensitive ($\alpha_\text{A}=0.5$).} This parameters regulates the expansion of the selected token set via attention mechanisms. Higher values yield tighter constraints (smaller editing areas), while lower values allow for broader structural changes. Similar to $T_\text{repaint}$, this is \textit{intention-dependent}. For example, replacing an entire head requires a higher $\alpha_\text{A}$, whereas swapping only facial details requires a lower $\alpha_\text{A}$. The qualitative example can be found in \cref{fig:ablation_gating}

\subsection{Texturing Process}
\noindent\textbf{Texture Difference Threshold ($\tau_\text{texture}=0.005$).} This controls the sensitivity of the preservation mask. A larger value preserves more of the original texture but may lead to misalignment with the condition $I_E$. A lower value allows the condition image to influence a larger portion of the surface, ensuring better visual alignment at the cost of original texture preservation.

\noindent\textbf{MV-RePaint Timestep ($T_\text{MV-repaint}=1.0$).} This controls the noise level for texture generation. We use the maximum value to ensure full denoising, guaranteeing global coherence in the generated texture.

\section{Limitations}
\noindent\textbf{Hyperparameter Sensitivity.} 
Although each proposed module serves a distinct role in the editing pipeline, disentangling their effects remains challenging. Unlike end-to-end methods controlled solely by text instructions, our approach involves multiple thresholds (e.g., for token selection and pruning) that require careful tuning to balance editing strength with preservation.

\smallskip

\noindent\textbf{Texture-Geometry Dependency.}
While our \textit{Detail-preserving Texture Baking} effectively retains the original textural information, it relies on the accurate localization of edits. In cases where the editing process inadvertently alters the geometry of the preserved region (i.e., mask leakage), the texture projection may become misaligned or produce artifacts.
\begin{figure*}
\includegraphics[width=\linewidth]{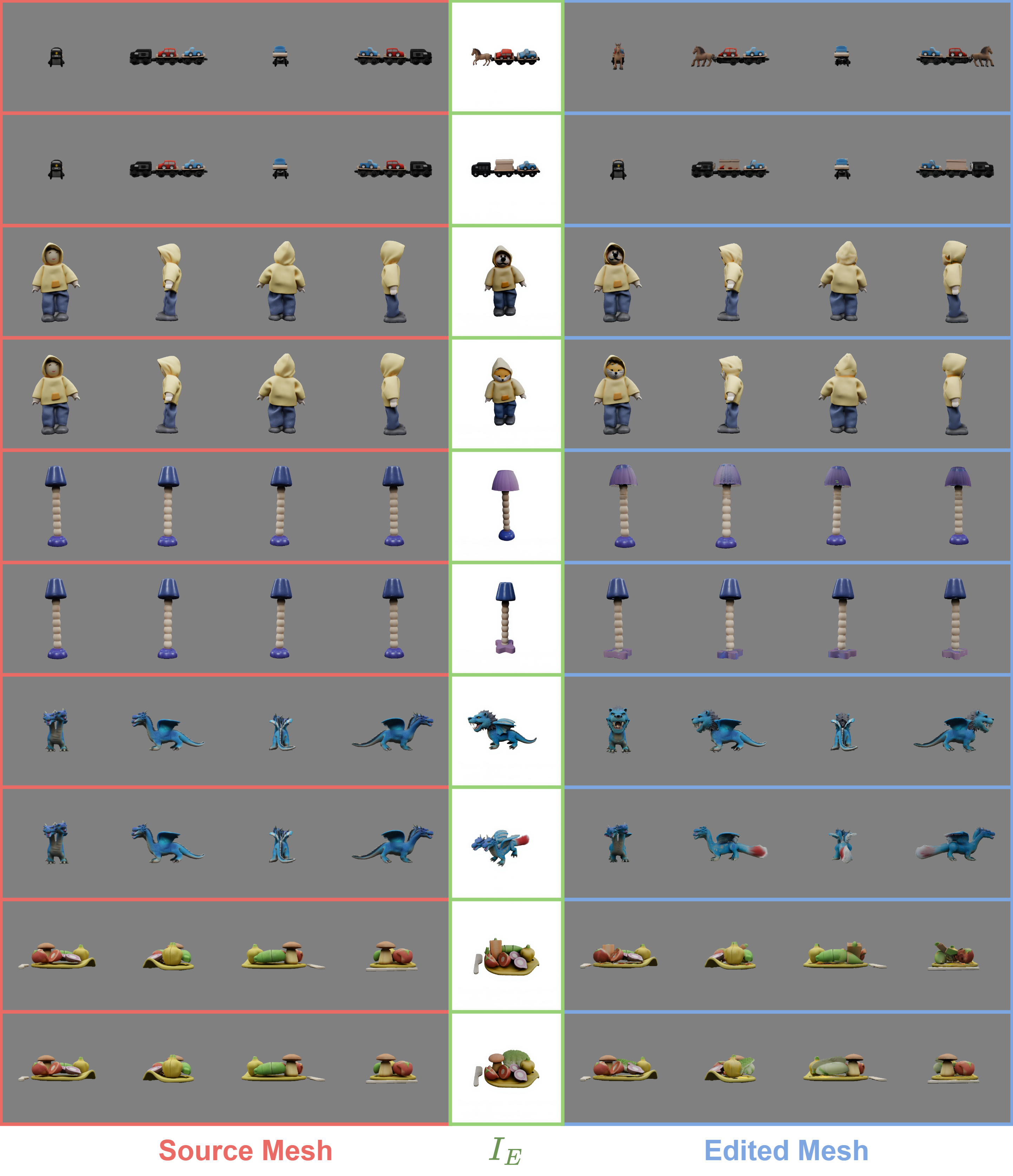}
\caption{More qualitative demonstration.}
\end{figure*}
\begin{figure*}
\includegraphics[width=\linewidth]{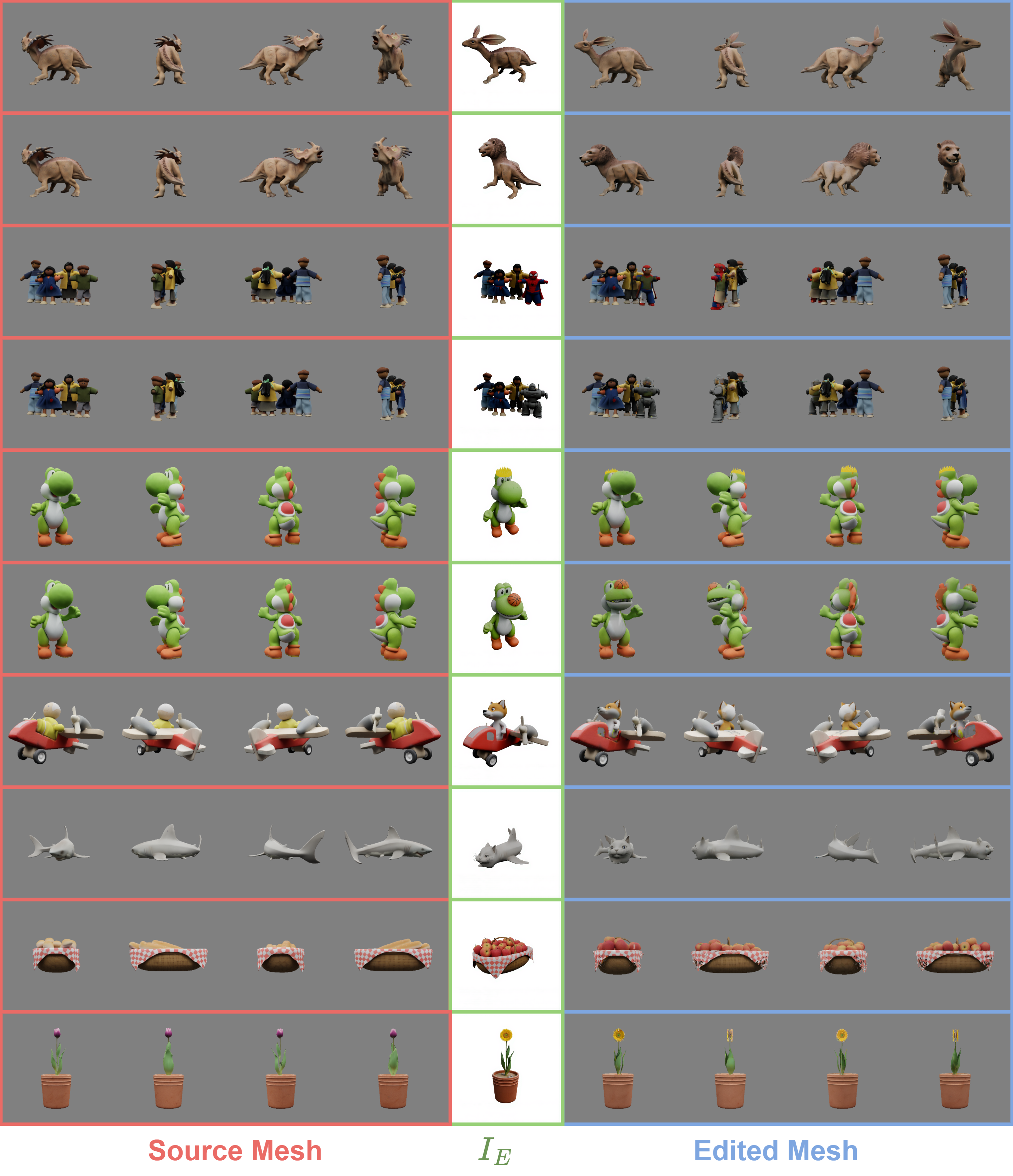}
\caption{More qualitative demonstration.}
\end{figure*}


\end{appendices}

\end{document}